\documentclass{article}

\pdfoutput=1
\usepackage{fullpage}
\usepackage[utf8]{inputenc}
\usepackage{graphicx}
\usepackage{amssymb}
\usepackage{amsmath}
\usepackage{amsfonts}
\usepackage{tabularx,colortbl}
\usepackage{multirow}
\usepackage{color}
\usepackage[pdftex]{hyperref}
\usepackage[english]{babel}
\usepackage{subfigure}
\usepackage{algorithm}
\usepackage{algorithmic}
\usepackage[sort&compress,numbers]{natbib}
\usepackage{authblk}

\definecolor{lgray}{rgb}{0.92,0.92,0.92}

\newcommand{\argmax}{\operatornamewithlimits{arg\ max}}

\makeindex

\begin{document}

\title{Entropic one-class classifiers}

\author[1]{Lorenzo Livi\thanks{llivi@scs.ryerson.ca}\thanks{Corresponding author}}
\author[1]{Alireza Sadeghian\thanks{asadeghi@ryerson.ca}}
\author[2]{Witold Pedrycz\thanks{wpedrycz@ualberta.ca}}
\affil[1]{Dept. of Computer Science, Ryerson University, 350 Victoria Street, Toronto, ON M5B 2K3, Canada}
\affil[2]{Dept. of Electrical and Computer Engineering, University of Alberta, Edmonton, AB T6G 2V4, Canada}
\renewcommand\Authands{, and }
\providecommand{\keywords}[1]{\textbf{\textit{Index terms---}} #1}

\maketitle

\begin{abstract}
The one-class classification problem is a well-known research endeavor in pattern recognition. The problem is also known under different names, such as outlier and novelty/anomaly detection. The core of the problem consists in modeling and recognizing patterns belonging only to a so-called target class.
All other patterns are termed non-target, and therefore they should be recognized as such.
In this paper, we propose a novel one-class classification system that is based on an interplay of different techniques.
Primarily, we follow a dissimilarity representation based approach; we embed the input data into the dissimilarity space by means of an appropriate parametric dissimilarity measure. This step allows us to process virtually any type of data.
The dissimilarity vectors are then represented through a weighted Euclidean graphs, which we use to (i) determine the entropy of the data distribution in the dissimilarity space, and at the same time (ii) derive effective decision regions that are modeled as clusters of vertices.
Since the dissimilarity measure for the input data is parametric, we optimize its parameters by means of a global optimization scheme, which considers both mesoscopic and structural characteristics of the data represented through the graphs.
The proposed one-class classifier is designed to provide both hard (Boolean) and soft decisions about the recognition of test patterns, allowing an accurate description of the classification process.
We evaluate the performance of the system on different benchmarking datasets, containing either feature-based or structured patterns.
Experimental results demonstrate the effectiveness of the proposed technique.\\
\keywords{One-class classification; Entropic spanning graph; Modularity measure; Dissimilarity representation; Fuzzy set.}
\end{abstract}

%%%%%%%%
\section{Introduction}

\begin{figure}[ht!]
 \centering
 \includegraphics[bb=0 0 448 437,scale=0.3,keepaspectratio=true]{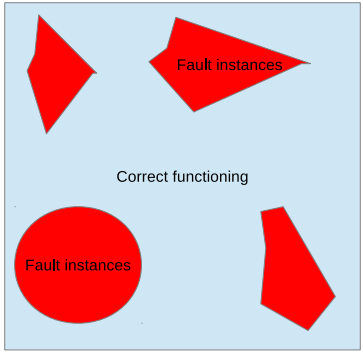}
 \label{fig:occ_example}
 \caption{2D space describing the status of a machine/device. Regions denoting instances of faults are depicted in red (colors online). The model of a OCC consists of those regions.}
\end{figure}

Pattern recognition problems involving the processing of patterns belonging to one class only are quite common \cite{Kemmler2013,Tax19991191,Juszczak20091859,occ_sg_enricods__arxiv,Tax:2002:UOG:944790.944809,fuzzy_occ,NIPS2002_2163,wang2013position,bodesheim1991divergence,6619277,6186735,4049825,pimentel2014review,Ding2014313,6722892,bicego2009soft}.
The interest in this type of problems is both methodological and of application-oriented character. In fact, one-class classification problems could be used to deal with tasks involving recognition of outliers in data. On the application side, instead, there are many real-world scenarios in which it is possible to obtain (or design) patterns only for the so-called ``target'' class. As an instance, we may cite the problem of determining whether a given machine/device is not working properly.
Intuitively, patterns representing the correct functioning of the machine/device are ``trivial'' and ``not informative'', in the sense that anything that is different from the observed faults is by definition an instance of correct functioning.
As a consequence, in this case one would model only those patterns representing fault instances (see Fig. \ref{fig:occ_example}).
However, modeling explicitly only one side of the decision boundary implies a more difficult setting with respect to (w.r.t.) well-established multi-class problems. In particular, the method for evaluating the performance of any one-class classifier (OCC) should take into account the implicit uncertainty rooted in the resulting decisions \cite{one-class_survey__2010,Juszczak20091859,fuzzy_occ,wang2013position,6722892}.

The one-class classification setting has been adopted in many real-world applications (for a review, see \cite{one-class_survey__2010,pimentel2014review}) such as the recognition of faults in smart electric grids \cite{occ_sg_enricods__arxiv}, Raman spectroscopy \cite{Kemmler201329}, events detection in videos \cite{piciarelli2008trajectory}, document classification \cite{Manevitz01one-classsvms}, medical imaging \cite{desir2012random}, and oil spill detection \cite{oilspill__2010}.

In this paper, we propose a novel OCC that is based on an interplay of different techniques. Our first objective is to make the proposed OCC applicable to any data type. To this end, we develop our approach on the basis of the dissimilarity space representation \cite{Duin2012826}. This choice, although increases the overall computational complexity, allows us to cover virtually any application context, regardless of the adopted representation for the input data (e.g., features, labeled graphs, etc.).
Then, we represent the embedded data in the dissimilarity space (DS) by a complete Euclidean graphs, whose vertices denote input patterns and the edges their mutual (normalized) Euclidean distance in the DS.
Such a graph allows us to (i) estimate the informativeness of the represented data and at the same time (ii) construct the decision regions (DRs) that we use to define the OCC model.
Additionally, representing the data in the DS by means of the Euclidean graph provides us a way to bypass all problems related to the high-dimensionality of the DS.
The informativeness of the embedded data is computed with the use of the $\alpha$-order R\'{e}nyi entropy, estimated by means of the entropic spanning graph technique proposed by \citet{gs:HeroEtAl2002}. In particular, we use the minimum spanning tree, which is further analyzed with the aim of inducing a partition of the graphs into suitably compact and separated clusters of vertices. We designed a fast graph partitioning algorithm based on the concept of modularity \cite{4358966}.
The derived DRs are then equipped with suitable membership functions \cite{Livi_ga_2013,rizzi2002}, which allow us to form both hard (Boolean) and soft decisions about the classification.
To benchmark the proposed OCC, we test two types of data: features and labeled graphs (also termed attributed graphs) based patterns.
We provide experiments and offer a comparative analysis for different datasets from well-known benchmarks, such as those coming from the UCI \cite{Bache+Lichman_2013} and IAM \cite{riesen+bunke2008} repositories.

The paper is structured as follows.
Section \ref{sec:background} offers an overview on the one-class classification context (Section \ref{sec:related_works}) by hence providing also a clear collocation for the proposed OCC system. Moreover, in Section \ref{sec:tech_background} we introduce the main technical background material used by the proposed OCC.
Successively, in Section \ref{sec:eocc} we present the details of the proposed OCC.
In Section \ref{sec:exps} we show and discuss the experimental evaluations.
Finally, in Section \ref{sec:conclusions} we draw our conclusions, providing also pointers to future directions.

%%%%%%%%
\section{Related Works and Background Material}
\label{sec:background}

\subsection{A Review on One-Class Classification Methods}
\label{sec:related_works}

\citet{pimentel2014review} group the current one-class methods in five different categories: (i) probabilistic, (ii) distance-based, (iii) domain-based, (iv) reconstruction-based, and finally (v) information-theoretic techniques.
Probabilistic methods are focused on the reconstruction of the generative probability density function (PDF) underlying the data at hand. Such methods are further subdivided into the usual parametric and non-parametric classes, where the former include methods based on the identification of the optimal parameters describing a pre-defined statistical model, while the latter include methods based on the reconstruction of the PDF directly from the data.
Distance-based methods operate, essentially, by means of a suitable distance measure in the input space. Techniques of this category can be grouped into clustering-based and nearest neighbors based approaches.
Reconstruction-based methods include classical data-driven approaches, such as neural networks and subspace-based methods.
Domain-based methods revolve around the well-established Support Vector Data Description (SVDD) \cite{Tax19991191}. In this case, the objective is to model the target data via suitable decision regions/surfaces by optimizing a specific convex optimization problem.
Finally, information-theoretic methods rely on information measures such as entropy, divergence, and mutual information.
Intuitively, a non-target pattern is identified as one that alters significantly the information content of the data.

As stated before, an important class of OCCs revolves around the SVDD method proposed by \citet{Tax19991191}, which has been elaborated taking inspiration from the well-known support vector machine (SVM) \cite{Tax19991191,SchWilSmoShaetal00,wang2013position}.
The classification model of SVDD is defined in terms of hyper-spheres, which cover the training set data through an SVM-like optimization problem (the minimization of the sphere radiuses is enforced).
SVDD is particularly exploitable since it can be used jointly with positive definite kernel functions, which allow the generalization of the input domain.
\citet{SchWilSmoShaetal00} proposed an alternative approach to SVDD that employs a hyperplane, like in the conventional SVM case.
The hyperplane is positioned to separated the region of the input space containing patterns form the region containing no data.
Other more recent approaches include algorithms based on the minimum spanning tree (MST) \cite{Juszczak20091859}, Gaussian processes \cite{Kemmler2013}, and on Random forests \cite{Desir20133490}.
The reader is referred to Ref. \cite{pimentel2014review} for a comprehensive survey portraying the state-of-the-art on one-class classification methods.

According to the aforementioned OCC systems categorization, the herein proposed OCC could be collocated in the intersection among probabilistic, distance-based, and information-theoretic based approaches.
Notably, the proposed OCC exhibits some linkages with the system of \citet{Juszczak20091859}, in the sense that their solution relies on a MST. However, their approach is substantially different, since they do not use either the information-theoretic, fuzzy sets, or graph partitioning concepts that we exploit in this study. Moreover, our approach is dissimilarity-based, which opens a way to a multitude of applications in different areas. This last aspect recalls the OCC scheme by \citet{NIPS2002_2163}; however the authors use different techniques to design their system, which are based on linear programming and prototype selection.
Graph-based, and in particular minimum spanning tree based, general clustering algorithms are popular in the literature \cite{10.1109/TPAMI.2012.226,1432700,xu2002clustering,Galluccio201396,Galluccio:2012:GBK:2184924.2185067}, since, in fact, a graph provides a powerful data abstraction and a sound mathematical framework. However, to the best of our knowledge, the use of graph-based entropy estimation techniques for the design of the OCC model is missing in the OCC literature.
The utilization of fuzzy sets to model the DRs establishes another connection with the so-called fuzzy one-class classifiers \cite{fuzzy_occ,6613141,Utkin:2012:FOC:2213741.2433967}.

\subsection{Technical Background Material}
\label{sec:tech_background}

In the following subsections, we introduce the main concepts used in the OCC proposed in this paper: dissimilarity representation, entropy estimation, and modularity of a graph partition.
For more detailed discussions on dissimilarity representation we refer the reader to \cite{pkekalska+duin2005}; for graph-based entropy estimation to \cite{intrdim_shapes_hero,gs:HeroEtAl2002}; finally, for modularity of a graph partition to \cite{fortunato2010,4358966}.

\subsubsection{Dissimilarity Representation}
\label{sec:dr}

In the dissimilarity representation \cite{pkekalska+duin2005,odse,Duin2012826}, the elements of an input dataset $\mathcal{S}\subset\mathcal{X}$ are characterized by considering their pairwise dissimilarity values.
The key component is the definition of a nonnegative (bounded) dissimilarity measure $d: \mathcal{X}\times\mathcal{X}\rightarrow\mathbb{R}^{+}$, which is in charge of synthesizing all relevant commonalities among the input patterns of $\mathcal{S}$ into a single real-valued number.
A set of prototypes, $\mathcal{R}$, called representation set (RS), is used to develop the dissimilarity matrix $\mathbf{D}$, which is given as $D_{ij}=d(x_i, r_j)$, for every $x_i\in\mathcal{S}$ and $r_j\in\mathcal{R}$. Of course, the determination of the most suitable $\mathcal{R}$, with usually $\mathcal{R}\subset\mathcal{S}$, is an important objective. Different techniques are discussed by \citet{pkekalska+duin2005}, which span from prototype selection strategies to criteria related to the embedded data.
By using directly the rows of $\mathbf{D}$ as embedding vectors, we can obtain the so-called dissimilarity space representation (DSR). This is the fastest way to represent the input data as $\mathbb{R}^d$ vectors by starting from the dissimilarity values. In addition, any common algebraic structure can be defined on the dissimilarity space (DS), making this approach very flexible.

An important property of the DSR is that distances in the DS are just scaled by a factor equal to $\sqrt{|\mathcal{R}|}$ w.r.t. those in the input space (see \cite[Sec.~4.4.1]{pkekalska+duin2005} for details).

\subsubsection{Graph-based Entropy Estimation}
\label{sec:e_mst}

Let $X$ be a continuous random variable with PDF $p(\cdot)$. The $\alpha$-order R\'{e}nyi entropy measure is defined as
\begin{equation}
\label{eq:differential_entropy}
H_{\alpha}(X)=\frac{1}{1-\alpha}\log\left(\int p(x)^{\alpha}dx\right), \ \alpha\geq0, \alpha\neq1.
\end{equation}

Let us assume to have a data sample $X_{n}$ of $n$ i.i.d. realizations of $X$, with $\underline{\mathbf{x}}_i\in X_{n}\subset\mathbb{R}^{d}, i=1, 2, ..., n$, and $d\geq 2$. Let $G$ be the complete Euclidean graph constructed over $X_{n}$.
An edge $e_{ij}$ connecting $\underline{\mathbf{x}}_i$ and $\underline{\mathbf{x}}_j$ is weighted using a weight based on their distance, $|e_{ij}| = d_{2}(\underline{\mathbf{x}}_i, \underline{\mathbf{x}}_j)$.
The $\alpha$-order R\'{e}nyi entropy (\ref{eq:differential_entropy}) can be estimated according to a geometric interpretation of an entropic spanning graph of $G$.
Examples of such graphs used in the literature are the MST, \textit{k}-NN graph, Steiner tree, and TSP graph \cite{md_ent,Bonev2013214,pal_renyi_e_knn__2010,odse2__arxiv,4897236,neemuchwala2005image,oubel2005assessment,intrdim_shapes_hero,Hero_Asympt__1999}. In this paper, we will focus on the MST \cite{bonev__2008,neemuchwala2005image}.
Let $L_{\gamma}(G)$ be the \textit{weighted length} of a MST, $T$, connecting the $n$ points in $X_{n}$,
\begin{equation}
\label{eq:l_mst}
L_{\gamma}(G) = \displaystyle\sum_{e_{ij}\in T} |e_{ij}|^{\gamma} ,
\end{equation}
where $\gamma\in(0, d)$ is a user-defined parameter.
The R\'{e}nyi entropy of order $\alpha\in(0, 1)$, elaborated using the MST length (\ref{eq:l_mst}), is defined as follows:
\begin{equation}
\label{eq:rentropy_mst}
\hat{H}_{\alpha}(G) = \frac{d}{\gamma}\left[ \ln\left(\frac{L_{\gamma}(G)}{n^{\alpha}}\right) - \ln\left(\beta(L_{\gamma}(G), d)\right) \right],
\end{equation}
where $\alpha=(d-\gamma)/d$.
The $\beta(L_{\gamma}(G), d)$ term is a constant that is defined as $\beta(L_{\gamma}(G), d) \simeq \gamma/2\ln\left( d/2\pi e \right)$.
As a consequence of $G$, the entropy estimator (\ref{eq:rentropy_mst}) is suitable for processing high-dimensional input data.

\subsubsection{Modularity of a Graph Partition}
\label{sec:modularity}

A graph $G=(\mathcal{V}, \mathcal{E})$ is a pair of vertices and edges. An edge models a binary relation among two vertices. Normally, an edge either exists or does not exist. However, in the case of \textit{weighted} graphs, every edge $e_{ij}\in\mathcal{E}$ is associated with a real-valued number called the weight, $w_{ij}=w(e_{ij})$, which determines the strength of the relation. The weighted adjacency matrix $\mathbf{A}$ of $G$ is defined as $A_{ij}=w_{ij}$.
The degree of a vertex $v_i\in\mathcal{V}$ is defined as $\mathrm{deg}(v_i)=\sum_{j=1}^{|\mathcal{V}|} A_{ij}$.
We will consider weighted graphs with weights in $[0, 1]$. If not diversely specified, when we refer to the ``number'' of edges we actually refer to the sum of their weights.

A partition \cite{Livi_ga_2013}, $K(G)$, of order $k$ of a graph $G=(\mathcal{V}, \mathcal{E})$, is commonly intended as a partition of the vertex set $\mathcal{V}(G)$ into disjoint subsets (clusters, modules), $K(G)=\{\mathcal{C}_1, \mathcal{C}_2, ..., \mathcal{C}_k\}$.
A well-established measure to determine the quality of $K(G)$ is the so-called modularity measure \cite{brandes+gaertler+wagner2003,fortunato2010,1367-2630-10-5-053039,PhysRevE.71.046117,4358966,rosvall2007information}, which basically quantifies how well $K(G)$ groups the vertices of $G$ into compact and separated clusters.
Intuitively, in a graph a cluster of vertices is compact if the number of the intra-cluster edges is considerably greater than the one of the inter-cluster edges.
The modularity measure $Q(\cdot, \cdot)$ is formally defined as follows,
\begin{align}
\label{eq:modularity}
&Q(G, K(G)) = \frac{1}{2|\mathcal{E}(G)|}\sum_{i=1,j=1}^{k} \left( A_{ij}-\frac{\mathrm{deg}(v_i)\mathrm{deg}(v_j)}{2|\mathcal{E}(G)|} \right) \delta(\mathcal{C}_i, \mathcal{C}_j),
\end{align}
where the graph $G$ is assumed to be partitioned according to a given $K(G)$.
Eq. \ref{eq:modularity} can be conveniently rewritten in terms of edges only (\cite{4358966,1367-2630-10-5-053039}):
\begin{equation}
\label{eq:modularity2}
Q(G, K(G)) = \sum_{l=1}^{k} \left[ \frac{|\mathcal{E}(\mathcal{C}_l)|}{|\mathcal{E}(G)|}-\left( \frac{\mathrm{deg}(\mathcal{C}_l)}{2|\mathcal{E}(G)|} \right)^2 \right].
\end{equation}

In the above expression, $|\mathcal{E}(\mathcal{C}_l)|$ is the number of intra-cluster edges, and $\mathrm{deg}(\mathcal{C}_l)$ is the sum of degrees of the vertices in the l\textit{th} cluster (considering all edges, i.e., also those with one end-point outside $\mathcal{C}_i$).
Detailing the terms in (\ref{eq:modularity2}), we have:
\begin{align}
|\mathcal{E}(\mathcal{C}_l)| = \sum_{\{e_{ij}|\ v_i,v_j\in\mathcal{C}_l\}} w_{ij};\ \ \ \mathrm{deg}(\mathcal{C}_l) &= \sum_{v_i\in\mathcal{C}_l} \mathrm{deg}(v_i); \ \ \  |\mathcal{E}(G)| = \sum_{e_{ij}\in\mathcal{E}(G)} w_{ij}.
\end{align}

The modularity of a graph $G$ is equal to the modularity of the partition of $G$ that maximizes (\ref{eq:modularity2}).
Finding such an optimal partition is NP-complete \cite{4358966}, and therefore many heuristics has been proposed in the literature \cite{fortunato2010}.
It is well-known that the modularity (\ref{eq:modularity2}) assumes values in $[-1/2, 1]$ \cite{4358966} (the higher, the better).
In this paper, we consider the normalized term $M(G, K(G))\in[0, 1]$, defined as:
\begin{equation}
\label{eq:normalized_modularity}
M(G, K(G))=\log(3/2 + Q(G, K(G)))/\log(5/2).
\end{equation}

%%%%%%%%
\section{The Proposed One-Class Classifier}
\label{sec:eocc}

Given an input dataset $\mathcal{S}\subset\mathcal{X}$, we design a one-class classifier that is applicable to any input domain, $\mathcal{X}$.
To fulfill such a requirement, we first embed the input data, $\mathcal{S}$, into an Euclidean space. Notably, we implement the embedding by constructing the DSR of $\mathcal{S}$ (see Sec. \ref{sec:dr}).
Let $d_{\mathrm{I}}: \mathcal{X}\times\mathcal{X}\rightarrow\mathbb{R}^+$ be a suitable dissimilarity measure.
We assume here that $d_{\mathrm{I}}(\cdot, \cdot)$ depends on some (numerical) parameters, say $p$, which alter the resulting view of the input data, $\mathcal{S}$, in the DS. The derived DSR of $\mathcal{S}$ is accordingly denoted with $D(\mathcal{S}, \mathcal{R}, p)$, where $\mathcal{R}\subseteq\mathcal{S}$ is the RS, and $p$ is the specific parameters instance of $d_{\mathrm{I}}(\cdot, \cdot)$.

The embedded data are then represented by constructing a complete graph $G=(\mathcal{V}, \mathcal{E})$.
Vertices $\mathcal{V}$ denote the patterns in $\mathcal{S}$, while edges denote their relations in terms of distance; each edge $e_{ij}\in\mathcal{E}$ has a weight given by $w_{ij}=d_2(v_i, v_j)$, where $d_2(\cdot, \cdot)$ is a suitable (normalized) Euclidean metric.
Graph representations are popular for high-dimensional patterns \cite{gs:HeroEtAl2002,oubel2005assessment}. Constructing $G$ allows us also to avoid the computational burden of determining the optimal RS, $\mathcal{R}$; the size of the RS corresponds to the dimensionality of the DS, and therefore usually a proper prototype selection method is used to consider the smallest but most informative subset of $\mathcal{R}$.
Since $d_2(\cdot, \cdot)$ is Euclidean, $G$ can be used also to estimate the $\alpha$-order R\'{e}nyi entropy \cite{intrdim_shapes_hero,gs:HeroEtAl2002} of the underlying data distribution through the computation of the entropic MST (see Sec. \ref{sec:e_mst}).
The entropy is a powerful mesoscopic data descriptor -- it is also a measure of the \textit{spread} of the data -- that we use together with other terms to guide the synthesis of the OCC model.
In the following, we denote with $G(p)$ the (complete) Euclidean graph constructed over $D(\mathcal{S}, \mathcal{R}, p)$, obtained by setting $d_{\mathrm{I}}(\cdot, \cdot)$ with the $p$ instances.

$G(p)$ provides an abstract framework in which we can develop the model, the related decision rules, and the synthesis of the OCC.
Modules (i.e., vertex clusters) of $G(p)$, which we denote as a collection $K(G(p))=\{\mathcal{C}_1, \mathcal{C}_2, ..., \mathcal{C}_k\}$, are considered as suitable DRs, i.e., the OCC model; we derive such DRs by exploiting the concept of modularity (see Sec. \ref{sec:modularity}).
We maximize the modularity (\ref{eq:modularity2}) of the graph by analyzing a partition derived directly on the entropic MST, which has to be computed in order to calculate the $\alpha$-order entropy.
The point here is that the vertex degrees in the MST are a scaled approximation of those in $G(p)$ -- highly central vertices in $G(p)$ remain well-connected also in the MST.
Let $T(G(p))$ be a MST of $G(p)$. By using $T(G(p))$, we can induce a partition $K(G(p))=\{\mathcal{C}_1, \mathcal{C}_2, ..., \mathcal{C}_k\}$ whose quality can be evaluated by considering all edges of $G(p)$.
Since the weights of $G(p)$ denote the pairwise normalized Euclidean distances in the DS, for the purpose of calculating the modularity (\ref{eq:modularity2}) of $K(G(p))$, we consider instead the quantity $\overline{w}_{ij}=1-w_{ij}$ (the lower the distance, the higher the contribution in terms of modularity).

In the one-class classification setting we model the target class only. Therefore, we need to conceive an inference mechanism that takes into account the implicit uncertainty in the definition of the DR boundaries.
To this aim, we equip each cluster $\mathcal{C}_i$ with a membership function \cite{pedrycz1990fuzzy,Livi_ga_2013,pedrycz1998introduction,kuncheva2000fuzzy}, which efficiently describes the uncertainty of the DR boundaries. In the following, we denote with $F(G(p))=\{\mathcal{F}_1, \mathcal{F}_2, ..., \mathcal{F}_k\}$ the fuzzy set-based DRs, which will be used during the test stage of the classifier to provide \textit{soft} decisions.

The aim of the synthesis is to optimize the OCC model by searching for the best $p$ instances. We provide two objective functions (related optimization problems are always intended to realize maximization).
The first one considers a linear convex combination of entropy and modularity, calculated on the training set $\mathcal{S}_{tr}\subset\mathcal{S}$ only; $\mathcal{S}_{tr}$ contains target patterns only.
The two terms are clearly in conflict. In fact, the entropy favors the general spread--separation of the data in the DS, while the modularity constraints the data to group into compact clusters of $G(p)$. This combination yields solutions that help magnifying the structure of the DRs in $G(p)$.
The second objective function is designed to train the OCC model by cross-validation, i.e., we effectively test the OCC model instances on a validation set, $\mathcal{S}_{vs}$, containing both target and non-target patterns.
The first approach is considerably faster for what concerns the training stage, although the second one provides a more effective solution in terms of test set recognition performance.

\begin{figure}[ht!]
 \centering
 \includegraphics[bb=0 0 571 407,scale=0.4,keepaspectratio=true]{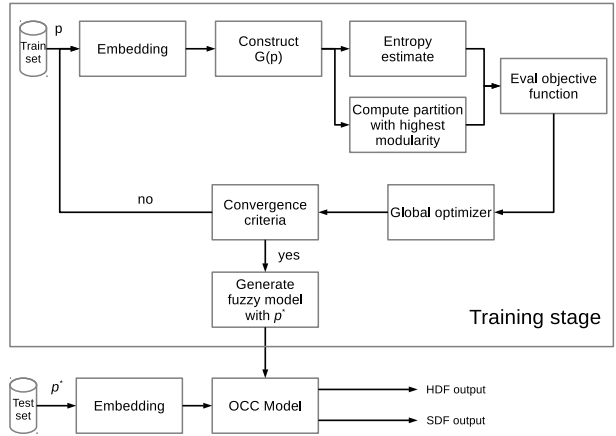}
 \caption{Block scheme of training and test stages of the OCC; first objective function is assumed. $\mathcal{S}_{tr}$ is used to synthesize the OCC model.
 The optimal parameters, $p^{*}$, are used to generate the fuzzy model, $F(G(p^{*}))$.
 $\mathcal{S}_{ts}$ is first embedded into the DS obtained by means of $p^{*}$, and successively it is tested. The OCC outputs both Boolean and membership grades to the target class.
}
 \label{fig:bloch_scheme}
\end{figure}

\begin{figure}[ht!]
\centering

\subfigure[Input data mapped into the DS.]{
\includegraphics[bb=0 0 535 267,scale=0.33,keepaspectratio=true]{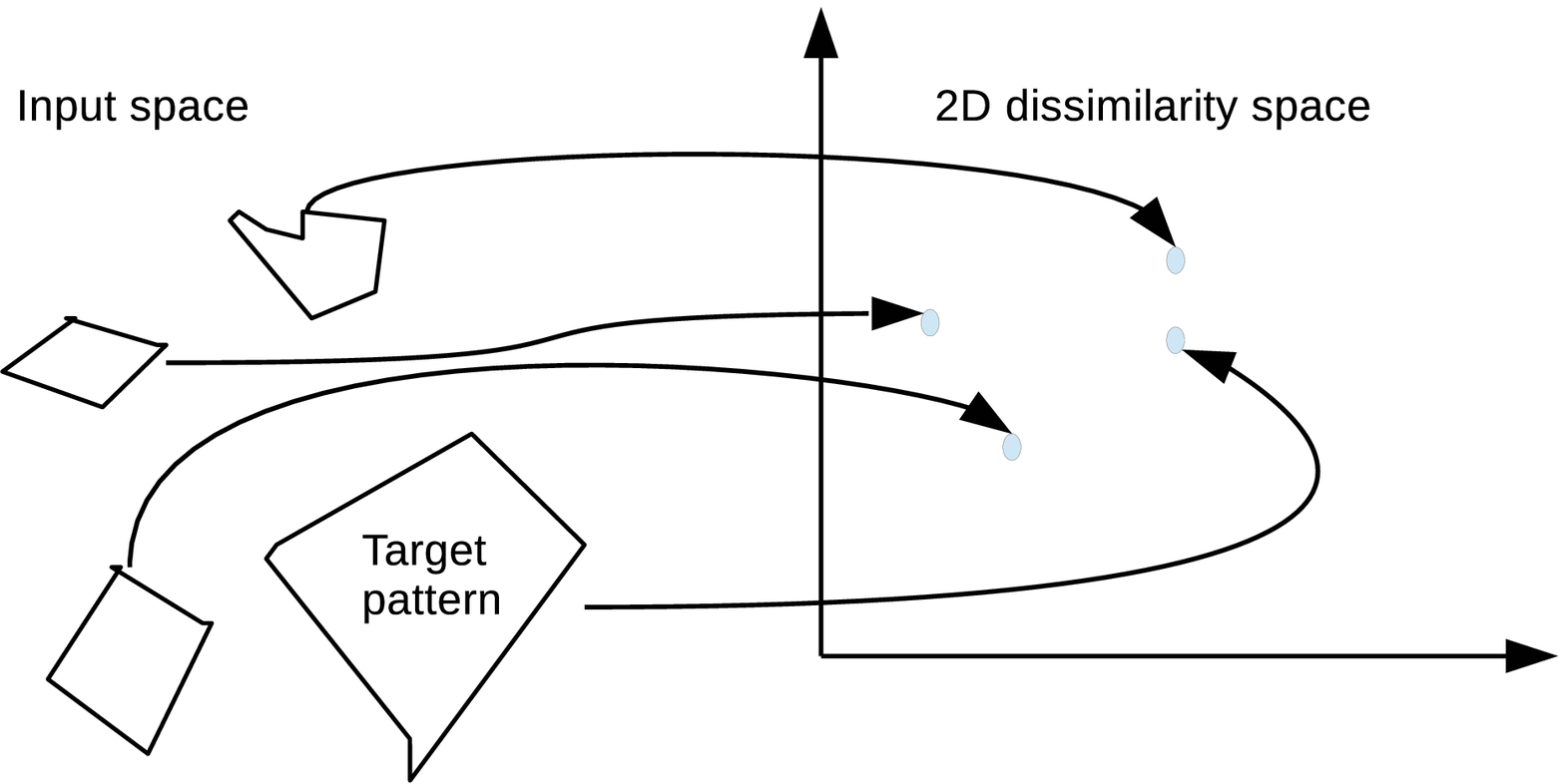}
\label{fig:functioning_1}}
~
\subfigure[Entropic spanning graph construction (MST).]{
\includegraphics[bb=0 0 570 230,scale=0.33,keepaspectratio=true]{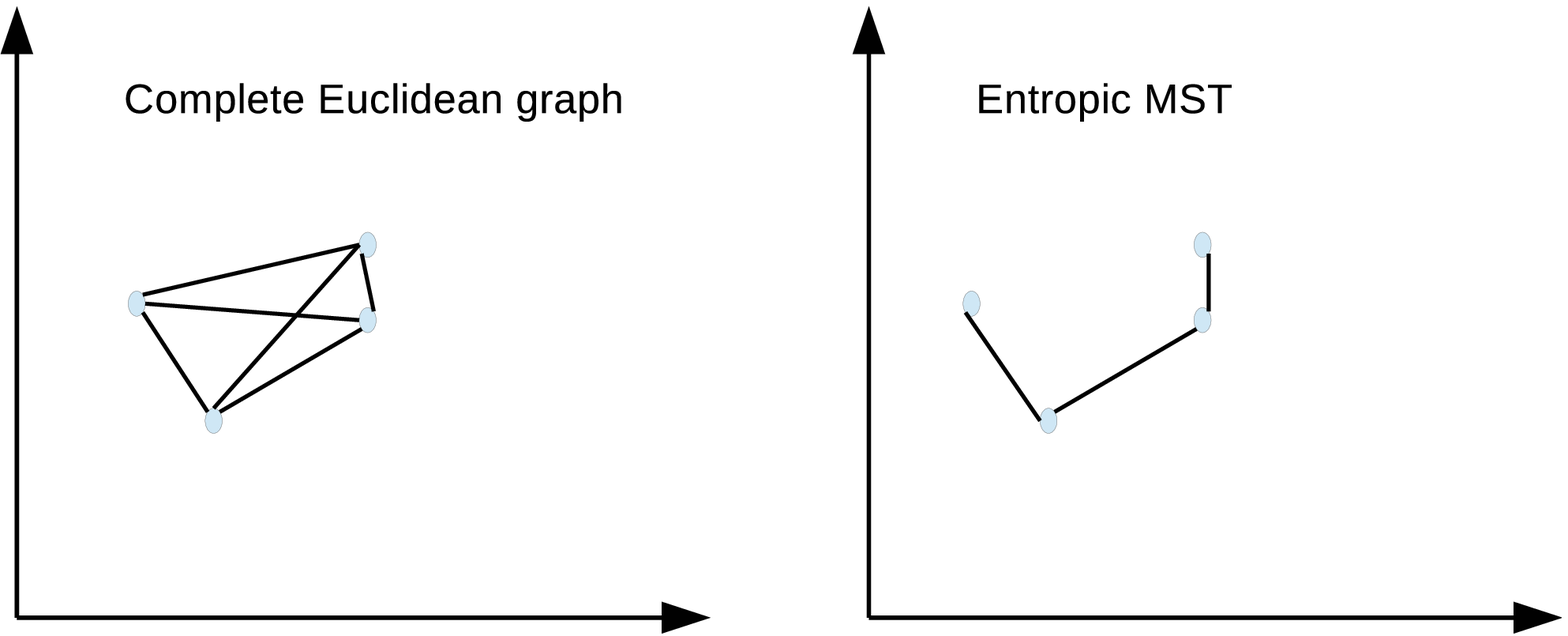}
\label{fig:functioning_2}}

\subfigure[Fuzzy model of the partition derived from the MST.]{
\includegraphics[bb=0 0 578 306,scale=0.33,keepaspectratio=true]{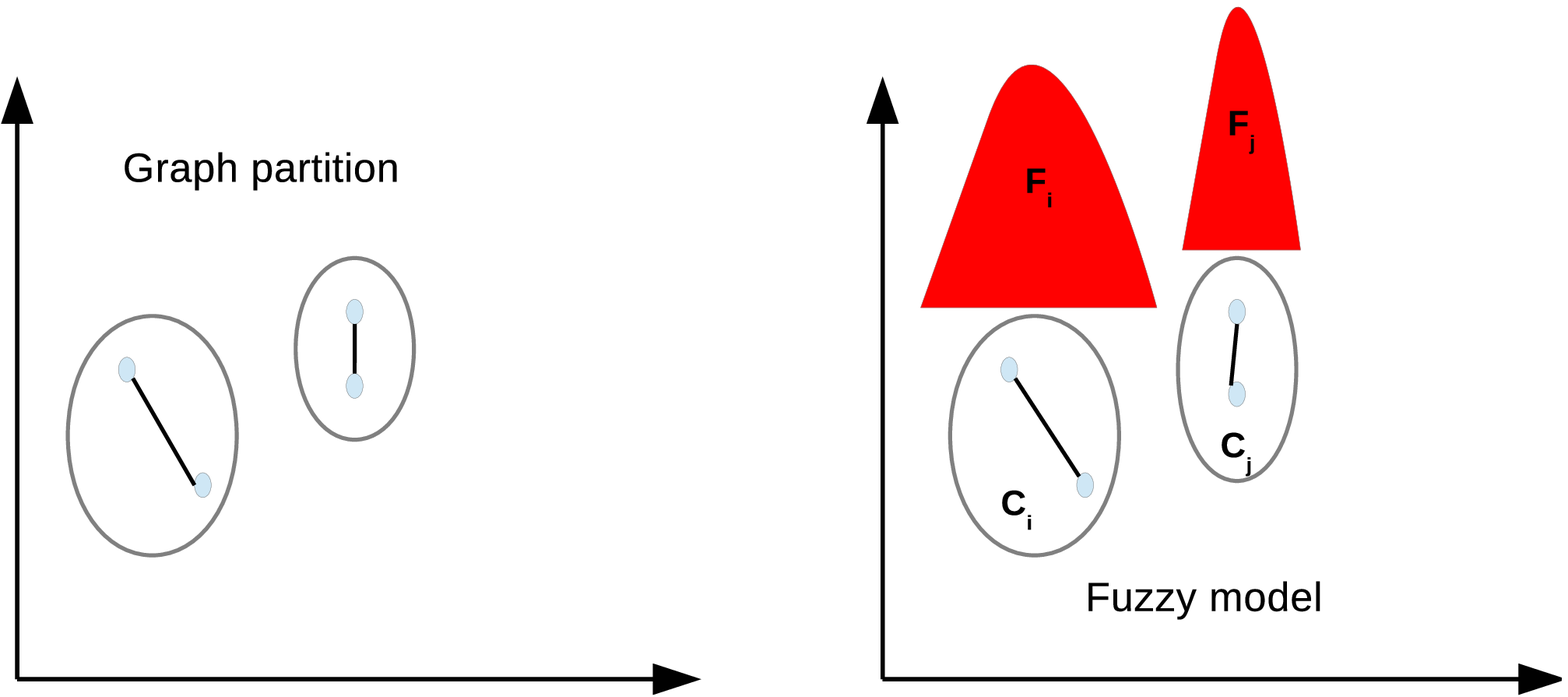}
\label{fig:functioning_3}}

\caption{Functioning of the proposed OCC; refer also to functional blocks in Fig. \ref{fig:bloch_scheme}.}
\label{fig:functioning}
\end{figure}

Fig. \ref{fig:bloch_scheme} provides a block scheme describing the proposed OCC, while Fig. \ref{fig:functioning} conveys the same information but using more intuitive illustrations.

\subsection{OCC Model and Related Testing}
\label{sec:occ_test}

A test pattern is classified by considering the optimal parameters $p^*$, which yield the graph $G(p^*)$ constructed over $D(\mathcal{S}_{tr}, \mathcal{R}, p^*)$, and the derived hard $K(G(p^*))$ and fuzzy $F(G(p^*))$ partitions.
Each fuzzy set $\mathcal{F}_i\in F(G(p^*)), i=1, 2, ..., k$, forms a fuzzy DR that is described by its membership function $\mu_{\mathcal{F}_{i}}(\cdot)$ and by a quantity $\tau_{i}>0$. The membership function $\mu_{\mathcal{F}_i}(\cdot)$ is parametrized by $\tau_{i}$, which is explicitly denoted as $\mu_{\mathcal{F}_{i}}(\cdot; \tau_i)$.
The scalar $\tau_{i}$ is determined by considering a statistics of the $\mathcal{C}_i$ intra-cluster edge weights (e.g., average, standard deviation).

A test pattern $x\in\mathcal{S}_{ts}$ is first mapped to a dissimilarity vector $\underline{\mathbf{v}}$ by setting $d_{\mathrm{I}}(\cdot, \cdot)$ with $p^*$ and considering the dissimilarity w.r.t. $\mathcal{R}$.
We define the soft decision function (SDF), which outputs a continuous value in $[0, 1]$ quantifying the membership degree of a test pattern to the target class, as:
\begin{align}
\label{eq:soft_decision_func}
\mathrm{SDF}(\underline{\mathbf{v}}) = \displaystyle\bot_{i=1}^{k} \mu_{\mathcal{F}_{i}}(\underline{\mathbf{v}}; \tau_i).
\end{align}

In the above expression, $\bot$ is a t-conorn (e.g., the maximum) and $\mu_{\mathcal{F}_{i}}(\cdot; \tau_i)$ is the membership function synthesized during the training -- the membership is a function of the distance of $\underline{\mathbf{v}}$ w.r.t. the cluster representative, $R(\mathcal{C}_i)$. Please note that here we do not provide closed-form expressions for $\bot$ and $\mu_{\mathcal{F}_{i}}(\cdot; \tau_i)$, since those two factors are general and they can be implemented in different fashions.
Later in the experiments section we will specify the setting that we adopted in this study.

It is important to provide also a hard (Boolean) decision function (HDF) about the classification. To this end, we exploit the cluster extent, $\tau_i$, for defining the HDF. Let
\begin{equation}
\mathcal{F}_{\mathrm{max}} = \argmax_{\mathcal{F}_i\in F(G(p^*))} \mu_{\mathcal{F}_{i}}(\underline{\mathbf{v}}; \tau_i)
\end{equation}
be the fuzzy set in which $\underline{\mathbf{v}}$ achieves the maximum membership degree; $\mathcal{C}_{\mathrm{max}}$ is the corresponding non-fuzzy DR.
Then,
\begin{equation}
\label{eq:hard_decision_1}
\mathrm{HDF}(\underline{\mathbf{v}}) =
\begin{cases}
1 & \mathrm{if}\ d_{2}(\underline{\mathbf{v}}, R(\mathcal{C}_{\mathrm{max}}))\leq \tau_{\mathrm{max}}, \\
0 & \mathrm{otherwise.}
\end{cases}
\end{equation}

\begin{figure}[ht!]
 \centering
 \includegraphics[bb=0 0 417 294,scale=0.55,keepaspectratio=true]{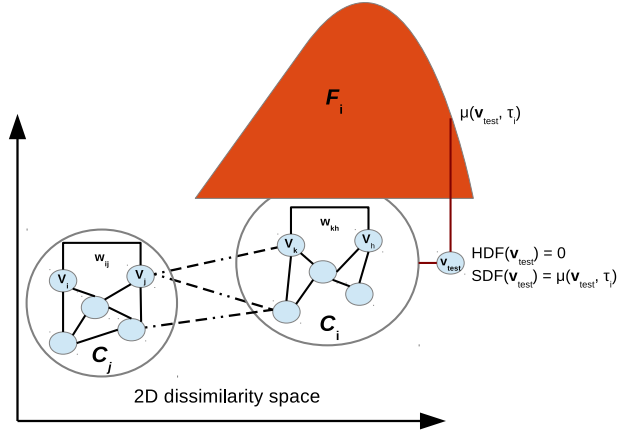}
 \caption{Fuzzy model constructed over $G(p)$. A test pattern, $\underline{\mathbf{v}}_{\mathrm{test}}$, is classified as non-target by the HDF, while considering the SDF it gets a membership degree, $\mu_{\mathcal{F}_i}(\underline{\mathbf{v}}_{\mathrm{test}}; \tau_i)$, to the target class.}
 \label{fig:fuzzy_model_graph}
\end{figure}

Fig. \ref{fig:fuzzy_model_graph} provides a schematic view of the test of an embedded pattern, $\underline{\mathbf{v}}_{\mathrm{test}}$.
Testing of the OCC model is characterized by a computational complexity given by the embedding of $x$ and the computation of both SDF and HDF.
The embedding can be performed in $O(|\mathcal{S}|+D |\mathcal{R}|)$, where $|\mathcal{S}|$ is the linear cost of deriving $\mathcal{R}$ from the input (training) data (that might be constant in the case $\mathcal{R}=\mathcal{S}$), and $D$ denotes the computational complexity of the dissimilarity measure for the input data. SDF can be computed by considering the $k$ different membership degrees; the same holds for HDF.

\subsection{Synthesis of the OCC Model}
\label{sec:fast_synthesis}

We cast the synthesis of the proposed OCC as the optimization of $G(p)$ w.r.t. the parameters $p\in\mathcal{P}$ of $d_{\mathrm{I}}(\cdot, \cdot)$. Note that in this paper, we make the fair assumption that $\mathcal{P}=[0, 1]^u$, where $u$ is the number of parameters characterizing $d_{\mathrm{I}}(\cdot, \cdot)$.
The idea is to determine the best-performing parameters setting so that the following objective function is maximized:
\begin{equation}
\label{eq:obj_func}
\max_{p\in\mathcal{P}} \eta \hat{H}_{\alpha}(G(p)) + (1-\eta) M(G(p), K(G(p))),\ \eta\in[0, 1].
\end{equation}

We evaluate two conflicting quantities on $G(p)$: the entropy, $H(G(p))$, and the modularity, $M(G(p), K(G(p)))$.
The entropy term favors the construction of a DS and related graph, $G(p)$, such that the overall distance among the vertices/patterns is maximized (the higher the entropy, the higher the spread of the data). On the other hand, evaluating the modularity of the derived partitioning, $K(G(p))$, constrains the optimization to search for solutions that magnify also the ``community'' structure of the graph -- community is a term \cite{fortunato2010} that is used to denote a compact and populated cluster of vertices.
This results in a graph $G(p)$ that is suitably optimized to derive the DRs in terms of compact and separated clusters of vertices.

However, to evaluate the modularity we first need to generate a partition, $K(G(p))$. In the next section, we describe the algorithm that we designed to derive a partition of $G(p)$, whose aim is to quickly find a reasonable approximation of the optimal modularity of $G(p)$.

\subsubsection{Greedy Edge Pruning Approach to Graph Partitioning}

The following algorithm exploits the fact that we need to compute the MST, $T(G(p))$, of $G(p)$ to calculate the $\alpha$-order R\'{e}nyi entropy (see Sec. \ref{sec:e_mst}).
Once we have the MST, we form clusters on $T(G(p))$ iteratively by pruning (i.e., removing) those edges with higher Euclidean distance values.
By construction of the MST, those edges are likely to be ``bridges'' among well-separated components of $G(p)$. A MST of a graph with $n$ vertices has $n-1$ edges, and hence the pruning loop is repeated at most $n-1$ times.
Therefore, at iteration $i=0, 1, ..., n-1$, we partition the vertices of the MST into $k=i+1$ connected components. The connected components of the MST are used to derive a partition on $G(p)$, by considering exactly the same grouping of the vertices. However, in $G(p)$ we have full information of the edges, which we use to compute the modularity of the resulting partition (\ref{eq:modularity2}).
To terminate the procedure, we exploit the following greedy assumption. Since the MST is connected, by first removing edges with maximum weight, we will form the most interesting communities/clusters in terms of modularity.
As a consequence, if at iteration $i+1$ we get a modularity value lower than the one obtained at iteration $i$, we stop the algorithm, returning the last computed partition.
Algorithm \ref{alg:edge_pruning} delivers the pseudo-code of the herein described procedure.
\begin{algorithm}[h!]\footnotesize
\caption{Greedy edge pruning algorithm.}
\label{alg:edge_pruning}
\begin{algorithmic}[1]
\REQUIRE The MST $T(G(p))$ and the graph $G(p)$ with $n$ vertices
\ENSURE A partition $K(G(p))$ of $G(p)$
\STATE Set $i=0$ and best modularity $M_i=-1$
\STATE Let $L$ be a list with the $n-1$ edges of $T(G(p))$ in non-increasing order
\FOR{$i=1, 2, ..., n-1$}
\STATE Remove $i$th edge from $T(G(p))$ and determine the resulting connected components, $K(T(G(p)))_i$
\STATE Derive the corresponding partitioning $K(G(p))_i$ of $G(p)$ by considering the same vertex grouping of $K(T(G(p)))_i$
\STATE Set $M_i$ according to the evaluation of (\ref{eq:modularity2}) on $K(G(p))_i$
\IF{$M_i<M_{i-1}$}
\RETURN $K(G(p))_{i-1}$
\ENDIF
\ENDFOR
\RETURN $K(G(p))_{i}$
\end{algorithmic}
\end{algorithm}

\subsubsection{Analysis of Computational Complexity}
\label{sec:cca_eocc1}

The overall computational overhead of synthesizing the OCC using the herein explained approach is characterized by the sum of the following costs: (i) embedding, (ii) graph construction and entropy estimation, (iii) determination of graph partition, and finally (iv) the generation of the membership functions. The first three components must be considered into a suitable optimization loop, while the last one is performed only once at the end of the optimization cycle.

The dissimilarity representation of the training set $\mathcal{S}_{tr}, n=|\mathcal{S}_{tr}|$, costs $O(n|\mathcal{R}|D)$, where $D$ is the cost of the dissimilarity measure for the input data. 
The second cost can be summarized as follows:
\begin{align}
\label{eq:rentropy_mst_complexity}
O\left( \frac{n(n-1)}{2}E + \frac{n(n-1)}{2}\times\log\left(\frac{n(n-1)}{2}\right) + (n-1) \right).
\end{align}

The first term standing in (\ref{eq:rentropy_mst_complexity}) accounts for the generation of $G(p)$, computing the respective Euclidean distances for the edge weights ($E$ is the cost).
The second term quantifies the cost involved in the MST computation using the well-known Kruskal's algorithm. The last term in (\ref{eq:rentropy_mst_complexity}) concerns the computation of the MST length.
The third cost is given by Algorithm \ref{alg:edge_pruning}.
The main cycle is repeated a maximum of $n-1$ times. At each iteration, we derive the connected components on $T(G(p))$, which costs $O(n-1)$. To induce the partitioning on $G(p)$, we simply cycle through the vertices, grouping them according to the connected components.
Eq. \ref{eq:modularity2} can be computed in $O(n(n-1))$. Putting all together, we have for Algorithm \ref{alg:edge_pruning}
\begin{equation}
O((n-1)\times( 2(n-1) + n(n-1) ),
\end{equation}
which is dominated by a cubic computational complexity in the number of training patterns.
The last cost (membership function elicitation) depends on the order of the derived best partition. In particular, for each cluster $\mathcal{C}_i$ we determine the membership function by first deriving $\tau_{\mathcal{C}_i}$. In the case average intra-cluster distances are considered, then the cost of this step is $O(2n)$.

The overall worst-case cost, considering as main parameter $n$, is given by the determination of the partition with best modularity. However, since we have designed the system to operate in the DS, the actual cost depends also on $D$, which may have a significant impact in case of complex input data types.

\subsection{Synthesis By Cross-validation}
\label{sec:synthesis_crossval}

Eq. \ref{eq:obj_func} defines an objective that does not take into account explicitly the recognition capability of the synthesized model.
A blind partitioning that derives $K(G(p))$ according to Algorithm \ref{alg:edge_pruning} might suffer from the problem of generating a too simple model, i.e., with too few DRs.
In fact, Algorithm \ref{alg:edge_pruning} constraints the partition to be formed only by those clusters that induce an well-defined community structure in $G(p)$. However, since in the one-class setting we synthesize the model on the target class only, a well-formed cluster/community structure in $G(p)$ may not be easy to identify, especially in hard problems.
By relaxing the imperative of finding the partition with maximum modularity, we can conceive another objective function that allows us to force the derivation of additional DRs.
The alternative objective function to be considered reads as,
\begin{align}
\label{eq:obj_func2}
\max_{p\in\mathcal{P}}\ &\beta P(\mathcal{S}_{tr}, \mathcal{S}_{vs}; F(G(p))) +& \\
\nonumber&(1-\beta) \left[\eta \hat{H}_{\alpha}(G(p)) + (1-\eta) M(G(p), K(G(p)))\right],
\end{align}
where $\eta,\beta\in[0, 1]$.
Eq. \ref{eq:obj_func2} takes explicitly into account a measure of recognition performance, $P(\mathcal{S}_{tr}, \mathcal{S}_{vs}; F(G(p)))$, achieved on a validation set, $\mathcal{S}_{vs}$.
This term is combined (again with a linear convex combination) with (\ref{eq:obj_func}). Therefore, the final model does not necessarily imply the best possible modularity, $M(G(p), K(G(p)))$, and entropy, $\hat{H}_{\alpha}(G(p))$, of $G(p)$, focusing instead on the solutions that perform better also in terms of recognition.
This choice, potentially, implies obtaining a more complex model (i.e., a partition characterized by more clusters, with lower overall modularity), although, as we will observe in the experiments, it usually provides also a more effective classification system on the test set.

Algorithm \ref{alg:crossval} shows the pseudo-code of the herein described training scheme.
DRs are derived by exploiting, basically, the same MST-based graph partitioning approach (Algorithm \ref{alg:edge_pruning}). In fact, DRs are derived incrementally, by first removing edges that are more likely to induce well-formed clusters (i.e., those edges with higher weights).
\begin{algorithm}[h!]\footnotesize
\caption{Training scheme by cross-validation.}
\label{alg:crossval}
\begin{algorithmic}[1]
\REQUIRE The training set $\mathcal{S}_{tr}$, the validation set $\mathcal{S}_{vs}$, and the parameters $p$
\ENSURE A fuzzy partition $F(G(p))$
\STATE Determine the DSR $D(\mathcal{S}_{tr}, \mathcal{R}, p)$ and $D(\mathcal{S}_{vs}, \mathcal{R}, p)$
\STATE Construct $G(p)$ over $D(\mathcal{S}_{tr}, \mathcal{R}, p)$
\STATE Estimate the entropy, $H(G(p))$, and get the MST, $T(G(p))$
\STATE Let $L$ be a list with the $n-1$ edges of $T(G(p))$ in non-increasing order
\STATE Set $P_{\mathrm{max}}=0$
\FOR{$i=1, 2, ..., n-1$}
\STATE Remove $i$th edge from $T(G(p))$. Determine the resulting connected components, $K(T(G(p)))_i$
\STATE Derive $K(G(p))_i$ of $G(p)$ by considering the same vertex grouping of $K(T(G(p)))_i$
\STATE Set $M_i$ according to the evaluation of (\ref{eq:modularity2}) on $K(G(p))_i$
\STATE Generate fuzzy model $F(G(p))_i$ from $K(G(p))_i$
\STATE Set $P_i$ as the evaluation of the objective function (\ref{eq:obj_func2}) on $\mathcal{S}_{vs}$
\IF{$P_i>P_{\mathrm{max}}$}
\STATE $P_{\mathrm{max}}=P_i$
\ENDIF
\ENDFOR
\RETURN $F(G(p))_{\mathrm{max}}$ corresponding to $P_{\mathrm{max}}$
\end{algorithmic}
\end{algorithm}

\subsubsection{Analysis of Computational Complexity}
\label{sec:cca_eocc2}

The training procedure described in Algorithm \ref{alg:crossval} implies a substantial change in the OCC scheme illustrated in Fig. \ref{fig:bloch_scheme}.
First, the modularity (\ref{eq:modularity2}) is computed always exactly $n-1$ times.
Additionally, the fuzzy model must be synthesized and then tested on $\mathcal{S}_{vs}$ inside the optimization cycle (see costs related to testing the model in Sec. \ref{sec:occ_test}). Those changes affect considerably the effective running time of the whole procedure, although the computational complexity, asymptotically, is not altered.

%%%%%%%
\section{Experimental Evaluation}
\label{sec:exps}

We start by describing the experimental setting and the considered performance measures (Sec. \ref{sec:pm_expsettin}).
Then, in Sec. \ref{sec:exp_synth} we provide some preliminary and explanatory tests on synthetically generated problems.
In Sec. \ref{sec:exp_uci} we perform experiments on different datasets of feature-based patterns taken from the UCI repository \cite{Bache+Lichman_2013}. Lastly, in Sec. \ref{sec:exp_iam}, we discuss the results obtained over different IAM datasets \cite{riesen+bunke2008}, containing patterns represented as labeled graphs.

\subsection{Experimental Setting and Performance Measures}
\label{sec:pm_expsettin}

All non-synthetic datasets considered in this paper are originally conceived for multi-class classification problems. Accordingly, we convert them to fit the one-class setting by selecting one class as the target class, and considering all other classes as non-target.
If not specified otherwise, we train the proposed OCC over the target patterns, $\mathcal{S}_{tr}$, while we test the model on all patterns, i.e., all targets not used during the training and all available non-target patterns.
In the following, we shorten the proposed system as EOCC (standing for Entropic One-Class Classifier). The variant operating with the training scheme described in Sec. \ref{sec:fast_synthesis} is denoted as EOCC-1, while EOCC-2 is used to refer to the variant operating as described in Sec. \ref{sec:synthesis_crossval}.

The global optimization is implemented by a genetic algorithm.
It performs roulette wheel selection, two-point crossover, and random mutation on the parameters characterizing the dissimilarity measure. In addition, the genetic algorithm implements an elitism strategy which automatically imports the fittest individual into the next population; we set the population size to 30 individuals and the mutation rate to 0.3. Convergence criteria is determined by combining a maximum number of iterations/evolutions (here set to 100) and a check that evaluates if the best fitness is not changed over the last ten evolutions. Such settings are determined by a preliminary tuning of the system.
Fuzzification of the vertex clusters is performed by generating Gaussian membership functions. The width/size of the Gaussian is set according to $\tau_{\mathcal{C}_i}$, which is computed as the intra-cluster average distance. Gaussian membership functions are symmetric and they can be described by a single parameter (i.e., the width/size). This fact is in agreement with the single-parameter description of the intra-cluster distances distribution.
The RS, $\mathcal{R}$, is defined equal to $\mathcal{S}_{tr}$ (no selection is performed at all). Note that performing the DSR of the input data is not necessary when processing feature-based patterns. However, we do embed also this type of data to provide a uniform view of the experimental results.
$\gamma$ in (\ref{eq:rentropy_mst}) is set to 0.02; $\eta,\beta$ in (\ref{eq:obj_func}) and (\ref{eq:obj_func2}) are both set to 0.5.
Again, those settings are determined by a preliminary fine-tuning stage.

Software is implemented in standard C++ by means of the SPARE library \cite{spare_graph_2013}. Tests have been executed on a machine running a 64-bit Linux OS, equipped with an Intel(R) Core(TM) i7-3930K CPU \@ 3.20GHz and 32 Gb of RAM.

When considering the SDF, we must rely on an appropriate test set performance measure that takes into account the ``scorings'' (in our case, the membership degrees to the target class) assigned to the test patterns.
In particular, in this paper we consider the Area Under the ROC Curve (AUC); AUC is computed according to Ref. \cite{Fawcett:2006:IRA:1159473.1159475}. AUC is a robust statistics that gives the average probability that a target pattern is ranked higher than a non-target one. 
When using the HDF, instead, we evaluate the obtained confusion matrix by analyzing standard measures such as accuracy, recall, precision, F-measure and so on.
The performance measure $P(\cdot, \cdot, \cdot)$ in (\ref{eq:obj_func2}) is defined as the accuracy on the validation set.

All results are reported in terms of averages of ten different runs executed with different random seeds; we report standard deviations and analysis of statistical significance with t-test.

\subsection{Synthetic Data}
\label{sec:exp_synth}

\begin{figure*}[ht!]
\centering

\subfigure[]{
\includegraphics[bb=0 0 341 241,scale=0.6,keepaspectratio=true]{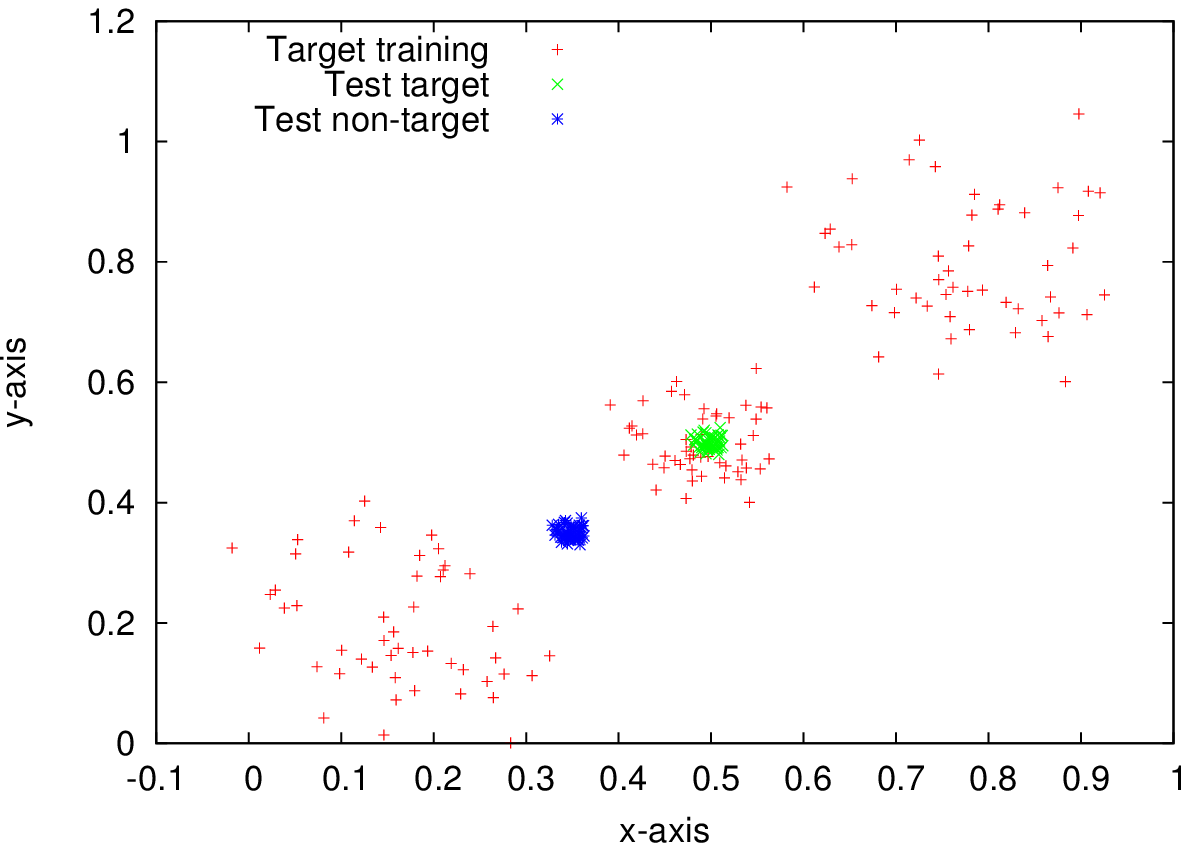}
\label{fig:ds1}}
~
\subfigure[]{
\includegraphics[bb=0 0 345 243,scale=0.6,keepaspectratio=true]{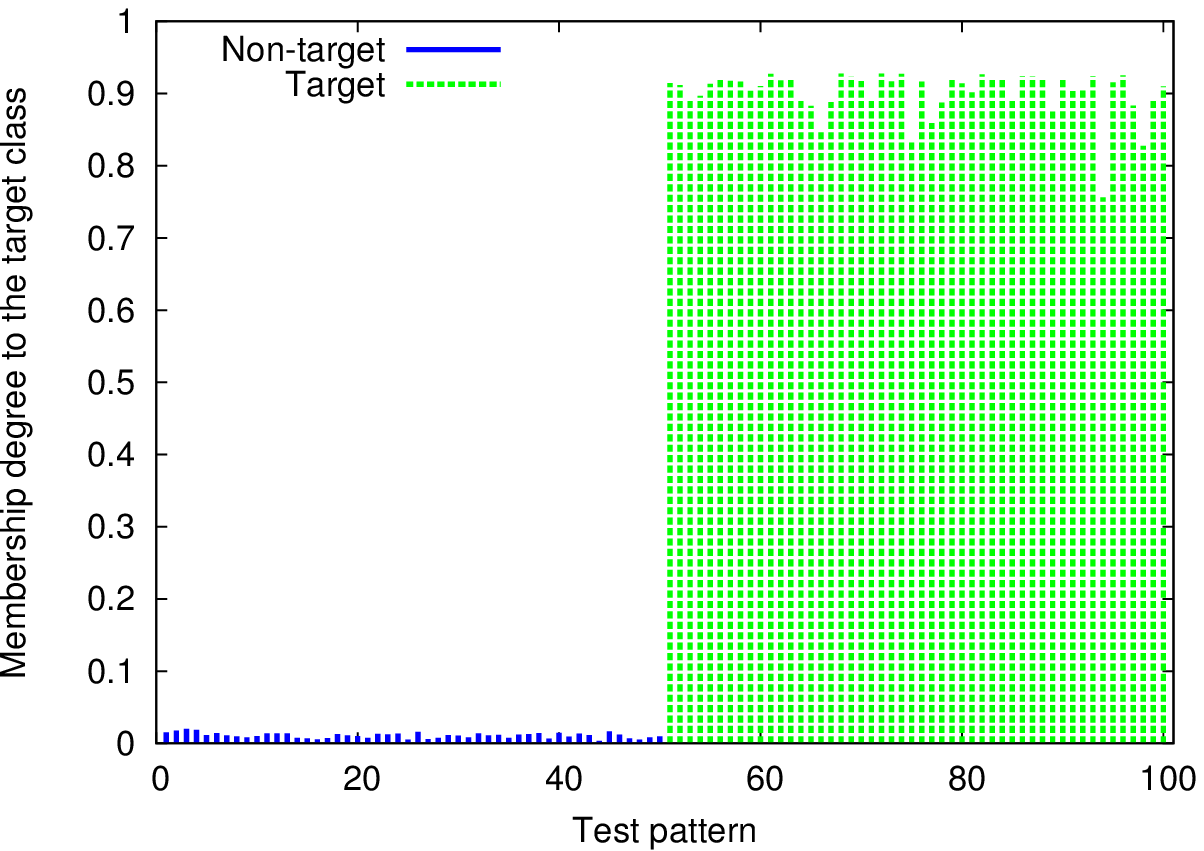}
\label{fig:ds1_memberships}}

\subfigure[]{
\includegraphics[bb=0 0 348 243,scale=0.6,keepaspectratio=true]{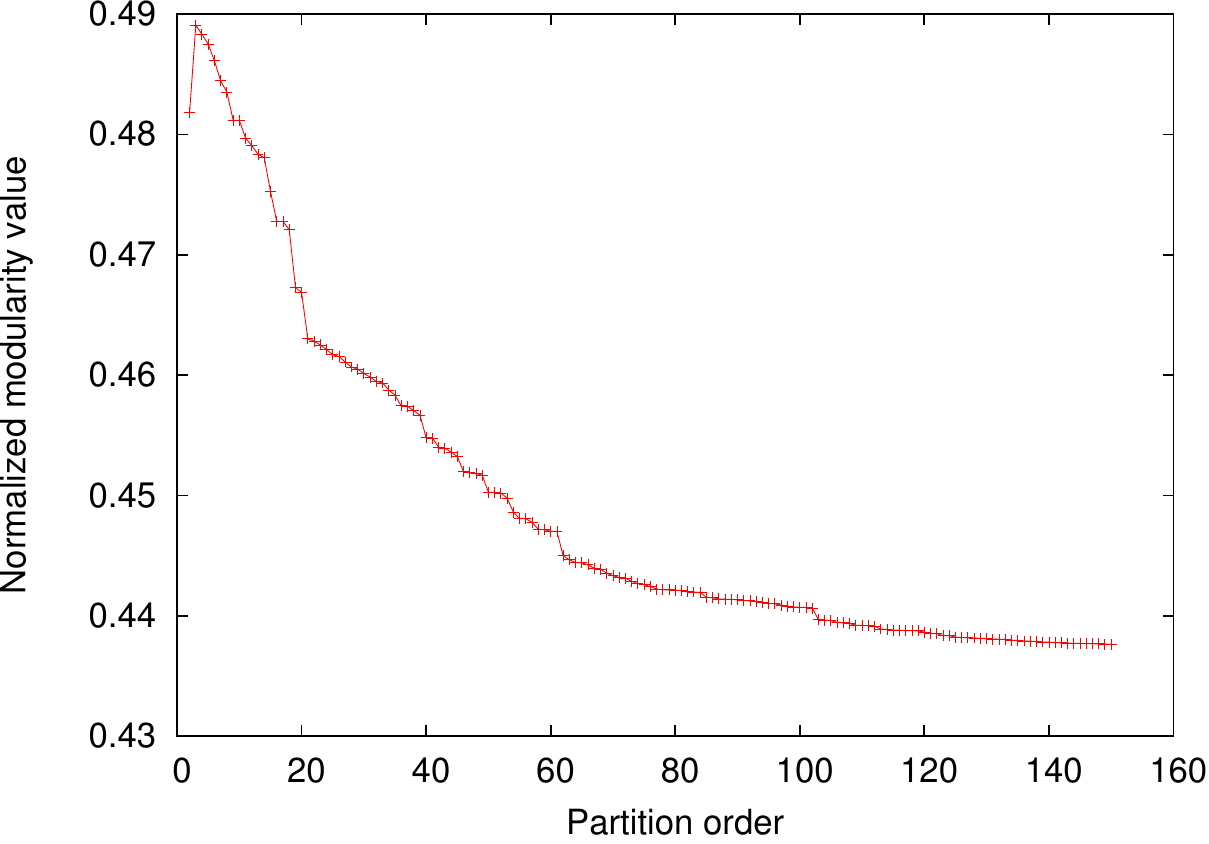}
\label{fig:ds1_modularity}}
~
\subfigure[]{
\includegraphics[bb=0 0 333 241,scale=0.6,keepaspectratio=true]{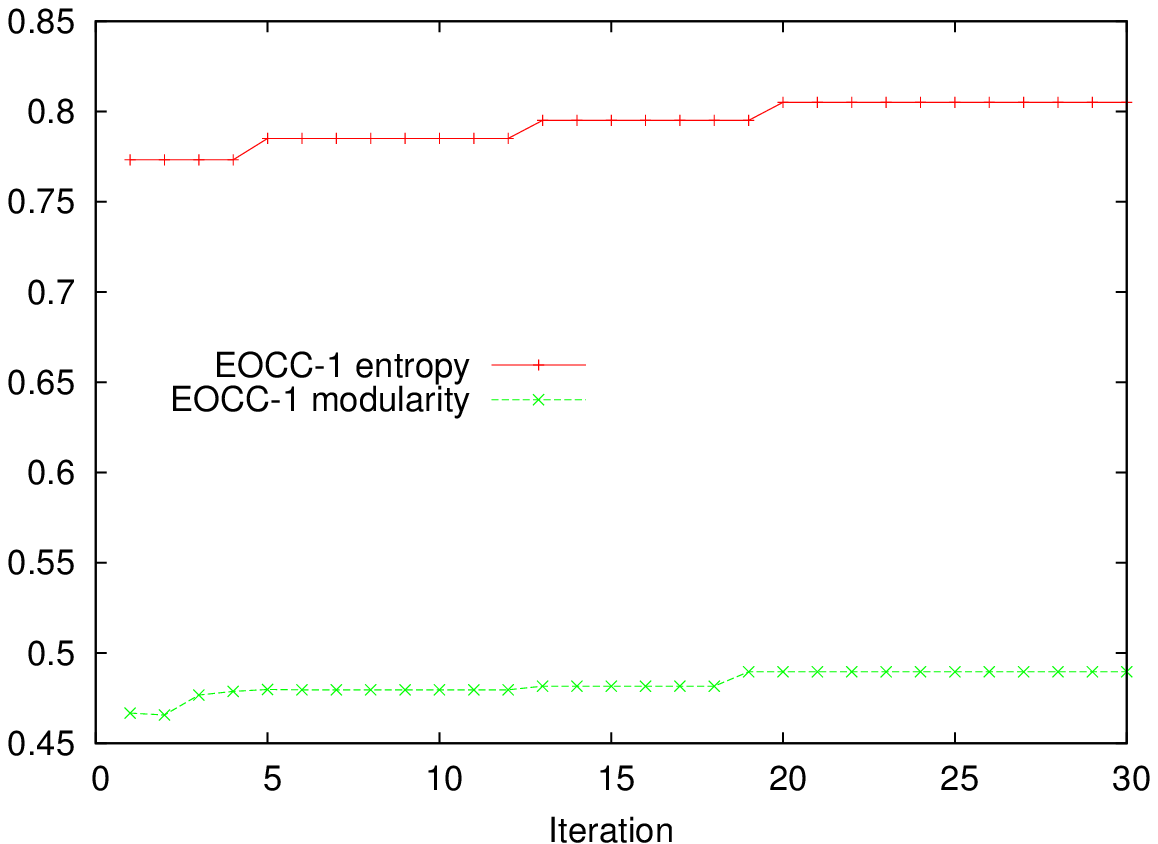}
\label{fig:ds1_entropy-modularity}}

\caption{In Fig. \ref{fig:ds1}, the red instances are used only during the training, while those in green are used for testing. Non-target instances are represented in blue. Fig. \ref{fig:ds1_memberships} shows the calculated membership values to the target class assigned by the SDF. Fig. \ref{fig:ds1_modularity} shows the trend of the modularity of all possible partitions derivable from the entropic MST. A peak is recognizable at the expected best partition order, i.e., three. Finally, Fig. \ref{fig:ds1_entropy-modularity} shows the entropy and modularity trends over the iterations of the optimization (EOCC-1 is assumed here).}
\label{fig:ds1_full}
\end{figure*}

Fig. \ref{fig:ds1_full} illustrates an example in which the target class is distributed in three well-separated spherical clusters. EOCC is trained on the target instances (in red) and it is tested on the green and blue instances (the green instances actually belong to the target class, while those in blue are non-target instances) -- see Fig \ref{fig:ds1}. This is a very simple instance and in fact the EOCC solves the problem without errors (considering both training schemes).
The obtained membership degrees to the target class are plotted in Fig. \ref{fig:ds1_memberships}.
The fact that EOCC synthesizes three DRs corresponds to the best solution in terms of modularity, as demonstrated in Fig. \ref{fig:ds1_modularity}, where the modularity value (\ref{eq:normalized_modularity}) of all possible partitions derivable from the MST are plotted -- note that here we consider a specific instance of the parameters of the input data dissimilarity measure (weighted Euclidean metric).
Finally, in Fig. \ref{fig:ds1_entropy-modularity} we show the entropy and modularity trends during the iterations of the optimization (\ref{eq:obj_func}). Since the problem is simple, the increments are numerically small, although the monotonic trend is clearly recognizable. It is worth noting the early convergence at the 30-th iteration, since in fact both entropy and modularity -- and hence the fitness (\ref{eq:obj_func}) -- are stuck at the same values.

\begin{figure*}[ht!]
\centering

\subfigure[]{
\includegraphics[bb=0 0 342 241,scale=0.6,keepaspectratio=true]{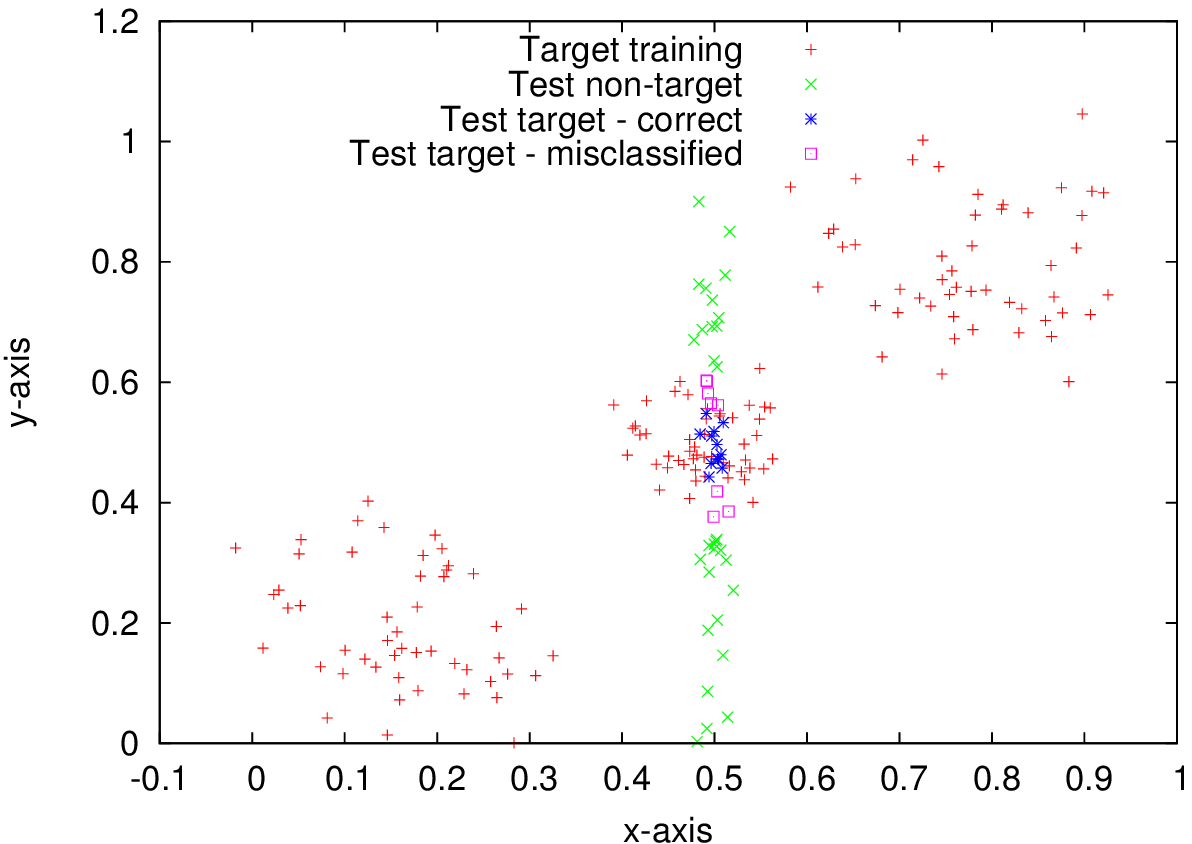}
\label{fig:ds2}}

\subfigure[]{
\includegraphics[bb=0 0 340 243,scale=0.6,keepaspectratio=true]{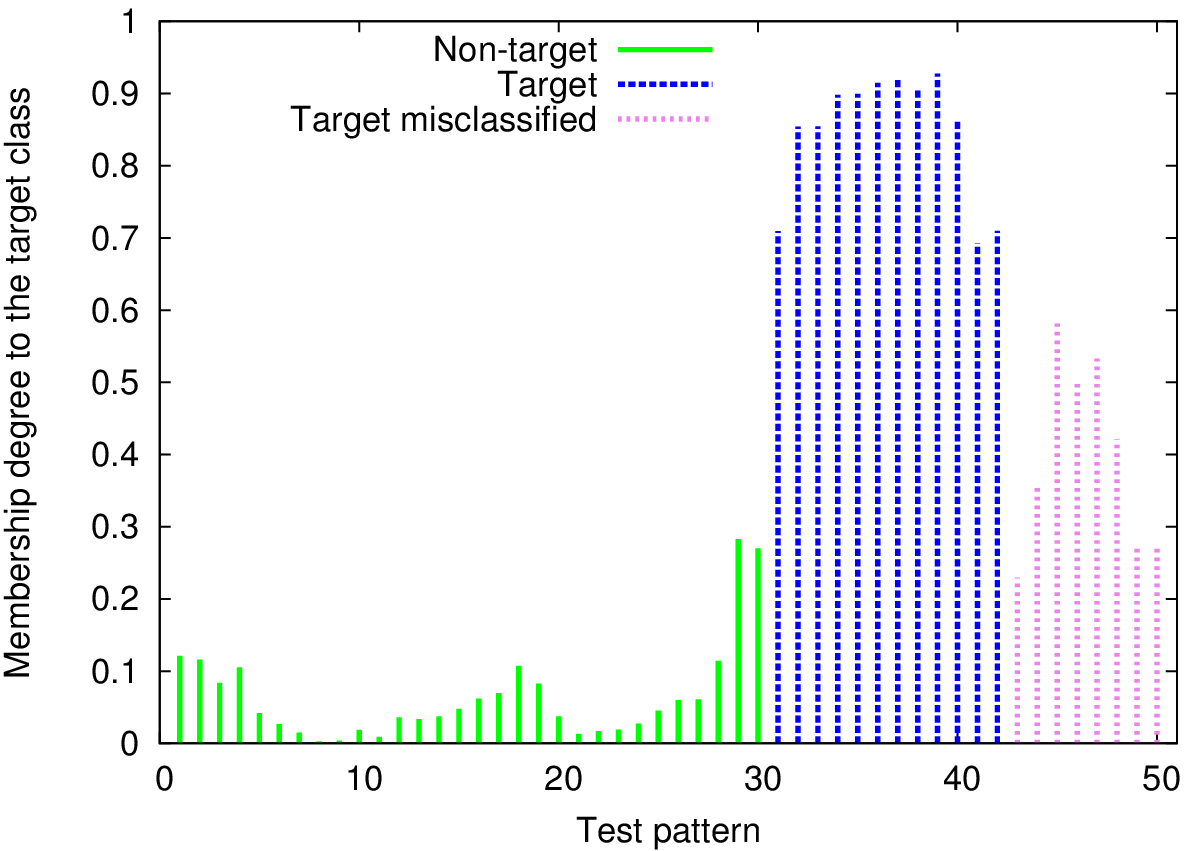}
\label{fig:ds2_memberships}}
~
\subfigure[]{
\includegraphics[bb=0 0 339 243,scale=0.6,keepaspectratio=true]{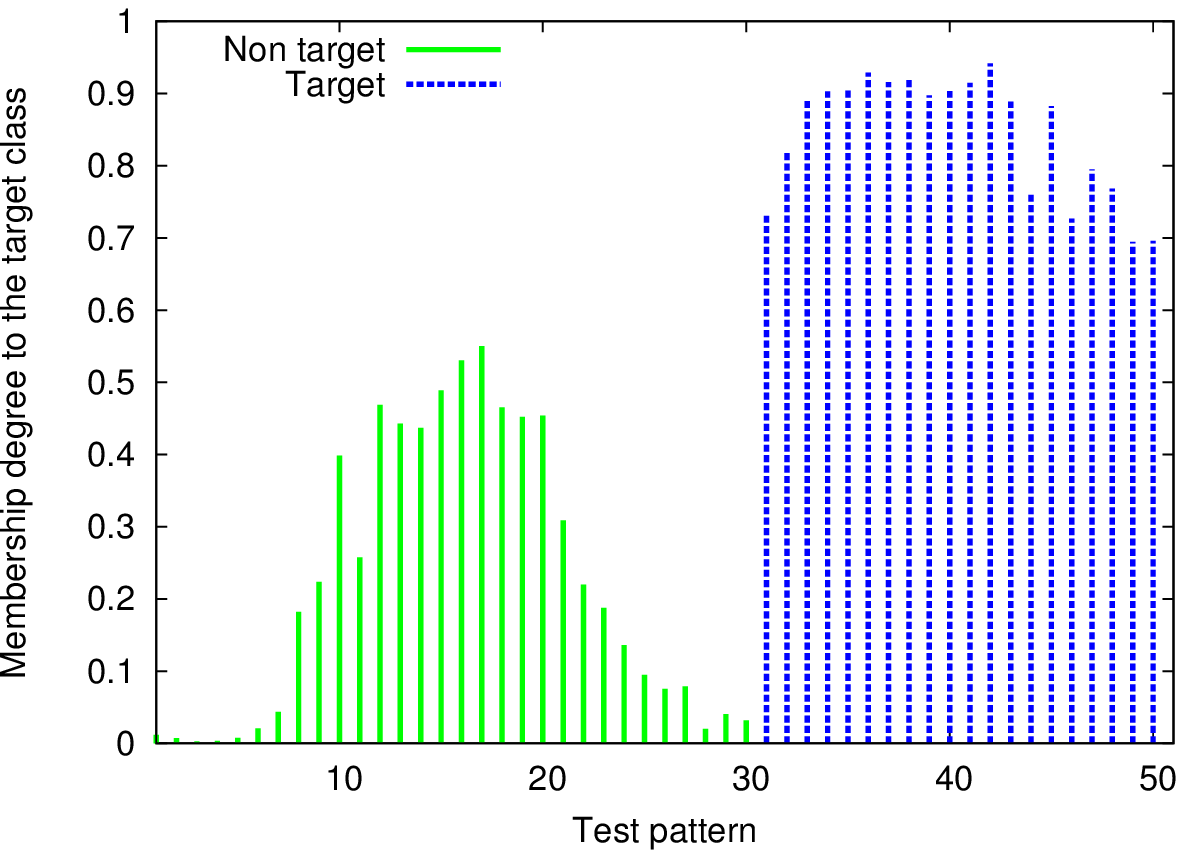}
\label{fig:ds2_memberships_crossval}}

\caption{In Fig. \ref{fig:ds2}, both blue and violet test patterns belong to the target class. The blue ones are correctly classified, while those in violet are misclassified in terms of HDF. Fig. \ref{fig:ds2_memberships} and \ref{fig:ds2_memberships_crossval} show, respectively, the membership degrees assigned to the test patterns by EOCC-1 and EOCC-2.}
\label{fig:ds2_full}
\end{figure*}

Fig. \ref{fig:ds2_full} shows a more subtle problem. The target instances shown in red are the same as in the previous problem. Pattern instances used for testing the OCC are now centered at $x=0.5$, but with a very narrow variance of the y-axis. Target instances used for testing are in blue and violet, while those represented in green are non-target instances -- see Fig. \ref{fig:ds2}.
Results achieved with EOCC-1 are good (for SDF, the AUC is 0.98), although it commits some errors considering the HDF (accuracy is 0.84, since eight test patterns of the target class are misclassified as non-target; represented in violet in figure). Three clusters are synthesized during the training. Fig. \ref{fig:ds2_memberships} reports the membership degrees obtained by EOCC-1.
On the other hand, EOCC-2 performs better in terms of SDF (see Fig. \ref{fig:ds2_memberships_crossval}; AUC is one, since EOCC-2 assigns always higher membership degrees to the target patterns. Moreover, also hard decisions are better, since zero errors are committed. Please note that, for clarity purpose, the validation data used for EOCC-2 is not shown in those examples.
This test is exemplar for the different views that can be obtained by considering soft or binary decisions in the one-class classification setting. In fact, the SDF may still provide a consistent picture of the correct labeling of the tested patterns even in complex situations. Additionally, this test yields a first insight on the possible difference in terms of recognition performance among EOCC-1 and EOCC-2 -- EOCC-2 is optimized explicitly towards those solutions that perform better.

\begin{figure*}[ht!]
\centering
\subfigure[]{
\includegraphics[bb=0 0 346 241,scale=0.6,keepaspectratio=true]{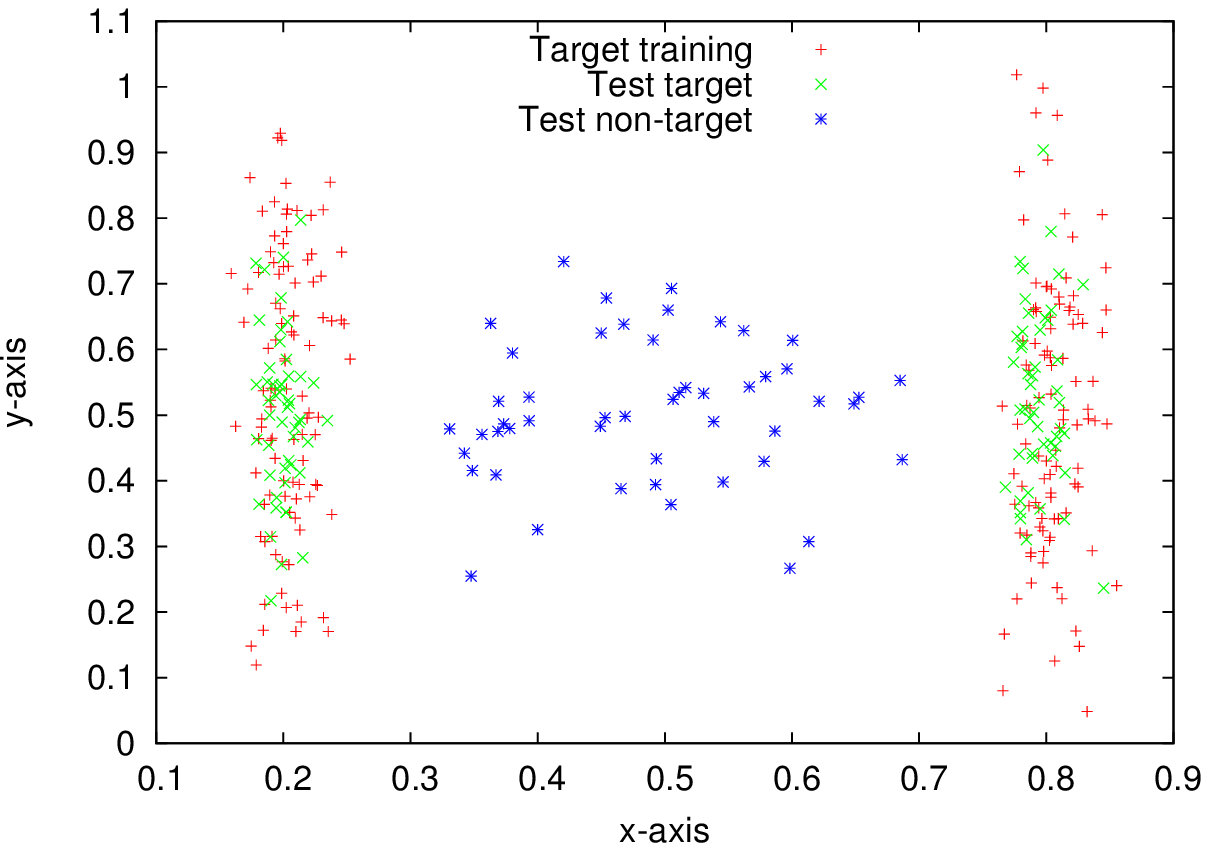}
\label{fig:ds3_ob}}
~
\subfigure[]{
\includegraphics[bb=0 0 341 241,scale=0.6,keepaspectratio=true]{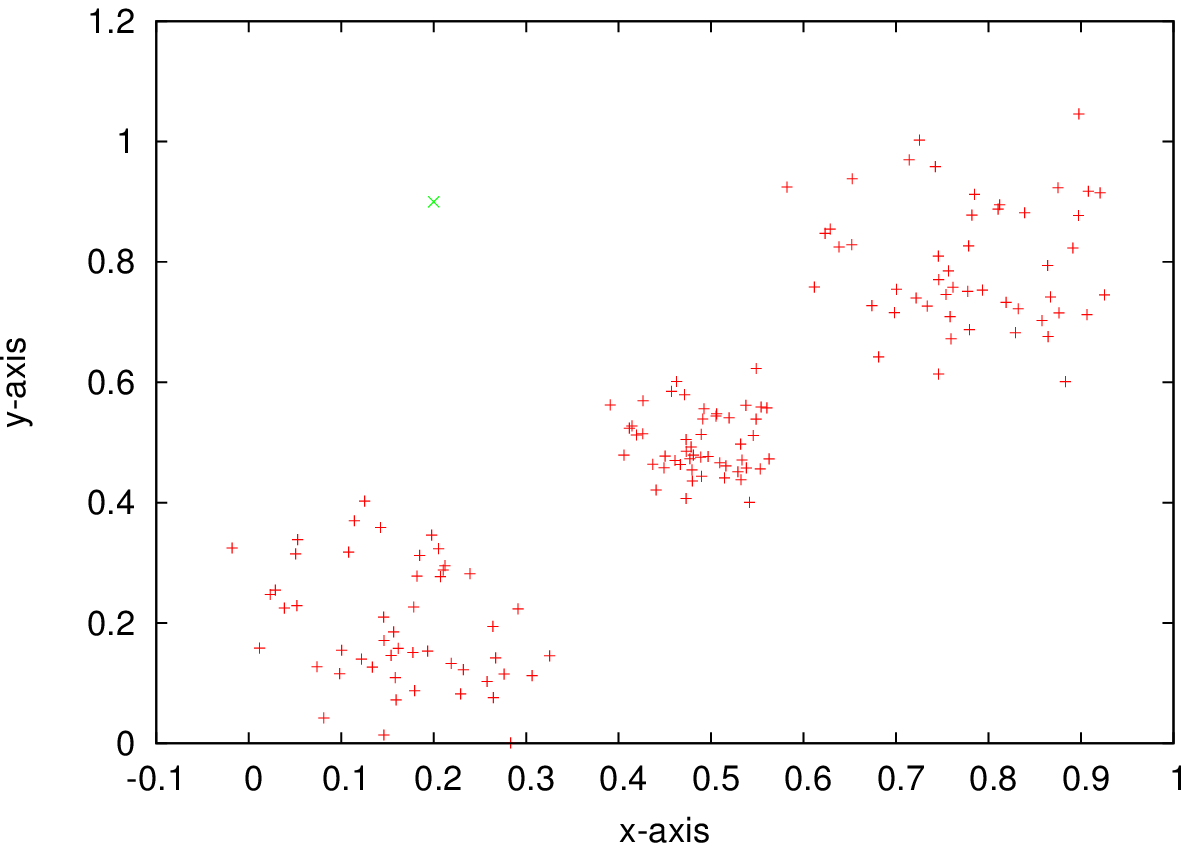}
\label{fig:ds3_isolated}}

\caption{Fig. \ref{fig:ds3_ob} depicts a situation where training/test target data shows high variance on the y-axis; the best-performing parameters, $p^*$, are selected accordingly. In Fig. \ref{fig:ds3_isolated} an example in which the training target data contains an isolated pattern (in green).}
\label{fig:ds3}
\end{figure*}

Finally, Fig. \ref{fig:ds3} shows two additional relevant properties of EOCC.
In Fig. \ref{fig:ds3_ob} the training/test target data are distributed in two highly separated clusters denoting high variance on the y-axis. By assuming the weighted Euclidean metric as dissimilarity measure, $p$ of (\ref{eq:obj_func}) would consist in a real-valued vector in $[0, 1]^2$. EOCC synthesizes a model with two DRs by finding the best-performing solution, $p^*$, equal to $[0.995, 0.238]$ (average of five runs).
This in fact corresponds to what we expected, since the x-coordinate plays the most important role in magnifying the separation (entropy), while the y-coordinate the compactness (modularity) of the partition.
In Fig. \ref{fig:ds3_isolated} we show a situation in which the training data contains an isolated pattern (shown in the upper-left corner in green). In our interpretation of the one-class classification setting, such a pattern is not an outlier, since in fact outliers should be identified during the test stage. As a consequence, EOCC synthesizes four DRs.

\subsection{Results on UCI Datasets}
\label{sec:exp_uci}

Tab. \ref{tab:uci_ds} presents the details of the herein considered UCI datasets \cite{Bache+Lichman_2013}.
\citet{Juszczak20091859} provide results of experimentations completed for the one-class setting, which we use for comparison in our study; where possible, missing results have been retrieved from \cite{OCC_results}. Please note that to provide a consistent comparison with Ref. \cite{Juszczak20091859}, we actually considered the versions of the UCI datasets downloaded from \cite{OCC_results}. We consider two versions of those data: (i) non-normalized (as downloaded from the reference) and (ii) normalized by ensuring zero-mean and unit variance for each component.
The UCI datasets usually do not provide a validation set explicitly. Since our principal aim here is the comparison of the proposed EOCC with the results presented in Ref. \cite{Juszczak20091859}, we need to consider the same training/test split scheme. As a consequence, in the case of EOCC-2 each validation set, $\mathcal{S}_{vs}$, is generated by applying a suitable zero-mean Gaussian noise to a randomly selected 10\% of the target/non-target data in $\mathcal{S}_{ts}$.
Nonetheless, as a demonstration of reliability of the method, we report also the results achieved defining a proper validation set by taking 10\% of the target and non-target patterns from $\mathcal{S}_{ts}$ (in the following tables, this is indicated as ``EOCC-2\_10\%'').
Since these datasets contain patterns described by features, the dissimilarity measure that we use is the weighted and normalized Euclidean metric, with weights $\underline{\mathbf{w}}\in[0, 1]^u$, where $u$ is the number of features characterizing the dataset at hand.
For settings of the herein considered reference systems we refer the reader to Ref. \cite{Juszczak20091859}.

Tab. \ref{tab:uci_ds_results} and \ref{tab:uci_ds_results2} show the average AUC results for EOCC-1, EOCC-2, and EOCC-2\_10\%, respectively on the considered low- and high-dimensional UCI datasets.
Results of Tab. \ref{tab:uci_ds_results} show very competitive performances of all EOCC variants w.r.t. the others; in four out of seven seven datasets (i.e., BW, D, E, and I) we achieve the highest AUC, considering either the non-normalized and the normalized dataset instances. Data normalization, usually, does not affect the results, with the only exception for the L dataset, in which EOCC performances degrades significantly; it is worth noting that this is observed also for the competitors.
Results of Tab. \ref{tab:uci_ds_results2} still denote good EOCC performances, although we score the best AUC in two datasets only (S in the non-normalized case, and in SP normalized setting). Results on the other datasets are in general comparable, although we observe performance degradation for all three systems on C; however, results are never statistically worse than all competitors.
Nonetheless, it is worth pointing out that, although EOCC relies on an entropy estimator (\ref{eq:rentropy_mst}) that in turn is (indirectly) based on the data PDF, we do not observe any severe performance breakdown when processing high-dimensional data, as it is usually observed with PDF-based methods (such as Parzen).
This follows from the fact that we initially perform a DSR, whose dimensionality is given by the size of the representation set. Therefore, since we used $\mathcal{R}=\mathcal{S}_{tr}$ in the experiments, one should expect to see some performance degradation as the training data size grows (for instance, when considering AB, BW, D, BC-D, and CO). However, on the contrary we achieve good results on those datasets, demonstrating the effectiveness of the combined dissimilarity and graph based approach adopted in EOCC.
Standard deviations are in general very low, denoting highly stable and reliable results.
\begin{table*}[th!]\scriptsize
\begin{center}
\caption{UCI datasets considered in this study.}
\label{tab:uci_ds}
\begin{tabular}{|c|c|c|c|c|c|c|}
\hline
\textbf{UCI Dataset} & \textbf{Acronym} & \textbf{Target class} & \textbf{\# Target} & \textbf{\# Non-target} & \textbf{\# Params} \\
\hline
\multicolumn{6}{|c|}{\textbf{\textit{Low-dimensional}}} \\
\hline
Abalone & AB & 1 & 1407 & 2770 & 10 \\
Biomed & BI & normal & 127 & 67 & 5 \\
Breast Wisconsin & BW & benign & 458 & 241 & 9 \\
Diabetes (prima indians) & D & present & 500 & 268 & 8 \\
Ecoli & E & pp & 52 & 284 & 7 \\
Iris & I & Iris-setosa & 50 & 100 & 4 \\
Liver & L & healthy & 200 & 145 & 6 \\
\hline
\multicolumn{6}{|c|}{\textbf{\textit{High-dimensional}}} \\
\hline
Arrhythmia & AR & normal & 237 & 183 & 278 \\
Breast cancer wisconsin (diagnostic) & BC-D & B & 357 & 212 & 30 \\
Breast cancer wisconsin (prognostic) & BC-P & N & 151 & 47 & 33 \\
Colon & C & normal & 22 & 40 & 1908 \\
Concordia & CO & 2 & 400 & 3600 & 1024 \\
Sonar & S & M & 111 & 97 & 60 \\
Spectf & SP & 0 & 95 & 254 & 44 \\
\hline
\end{tabular}
\end{center}
\end{table*}
\begin{table*}[thp!]\scriptsize
\begin{center}
\caption{Test set results on low-dimensional datasets. Results show the average AUC with standard deviations and significance test (t-test). Statistically significant results are in bold; not available is denoted with ``-''.}
\label{tab:uci_ds_results}
\begin{tabular}{|c|c|c|c|c|c|c|c|}
\hline
\textbf{System/Dataset} & \textbf{AB} & \textbf{BI} & \textbf{BW} & \textbf{D} & \textbf{E} & \textbf{I} & \textbf{L} \\
\hline
\multicolumn{8}{|c|}{\textbf{\textit{Non normalized data}}} \\
\hline
\rowcolor{lgray}EOCC-1 & 0.685(0.013) & 0.847(0.002) & \textbf{0.990(0.003)} & 0.607(0.046) & 0.953(0.002) & \textbf{1.000(0.000)} & 0.461(0.006) \\
\rowcolor{lgray}EOCC-2 & 0.831(0.001) & 0.864(0.003) & 0.989(0.001) & \textbf{0.717(0.005)} & \textbf{0.957(0.003)} & \textbf{1.000(0.000)} & 0.536(0.021) \\
\rowcolor{lgray}EOCC-2\_10\% & 0.819(0.002) & 0.868(0.007) & 0.929(0.061) & 0.677(0.019) & 0.954(0.003) & \textbf{1.000(0.000)} & 0.493(0.021) \\
\hline
Gauss & 0.861(0.002) & 0.900(0.004) & 0.823(0.002) & 0.705(0.003) & 0.929(0.003) & \textbf{1.000(0.000)} & 0.586(0.005) \\
MoG & 0.853(0.005) & 0.912(0.009) & 0.785(1.003) & 0.674(0.003) & 0.920(0.004) & \textbf{1.000(0.000)} & 0.607(0.006) \\
Na\"{\i}ve Parzen & 0.859(0.004) & \textbf{0.931(0.002)} & 0.965(0.004) & 0.679(0.003) & 0.930(0.008) & \textbf{1.000(0.000)} & \textbf{0.614(0.002)} \\
Parzen & 0.863(0.001) & 0.900(0.011) & 0.723(0.005) & 0.676(0.004) & 0.922(0.004) & \textbf{1.000(0.000)} & 0.590(0.003) \\
\textit{k}-Means & 0.792(0.011) & 0.878(0.012) & 0.846(0.035) & 0.659(0.007) & 0.891(1.006) & \textbf{1.000(0.000)} & 0.578(1.000) \\
1-NN & 0.865(0.001) & 0.891(0.008) & 0.694(0.006) & 0.667(0.007) & 0.902(0.009) & \textbf{1.000(0.000)} & 0.590(0.009) \\
\textit{k}-NN & 0.865(0.001) & 0.891(0.008) & 0.694(0.006) & 0.667(0.007) & 0.902(0.009) & \textbf{1.000(0.000)} & 0.590(0.009) \\
Auto-encoder & 0.826(0.003) & 0.856(0.022) & 0.384(0.009) & 0.598(1.008) & 0.878(1.000) & \textbf{1.000(0.000)} & 0.564(0.009) \\
PCA & 0.802(0.001) & 0.897(0.005) & 0.303(0.010) & 0.587(0.002) & 0.669(0.011) & 0.973(0.008) & 0.549(0.005) \\
SOM & 0.814(0.003) & 0.887(0.008) & 0.790(0.023) & 0.692(0.007) & 0.890(0.011) & \textbf{1.000(0.000)} & 0.596(0.007) \\
MST\_CD & \textbf{0.875(0.001)} & 0.898(0.010) & 0.765(0.018) & 0.669(0.007) & 0.897(0.009) & \textbf{1.000(0.000)} & 0.580(0.009) \\
\textit{k}-Centres & 0.760(0.008) & 0.878(0.024) & 0.715(0.124) & 0.606(0.016) & 0.863(0.012) & \textbf{1.000(0.000)} & 0.537(0.041) \\
SVDD & 0.806(0.001) & 0.220(0.003) & 0.700(0.006) & 0.577(0.098) & 0.894(0.008) & \textbf{1.000(0.000)} & 0.470(0.014) \\
MPM & 0.594(0.001) & 0.792(0.057) & 0.694(0.006) & 0.656(0.007) & 0.802(0.005) & \textbf{1.000(0.000)} & 0.587(0.009) \\
LPDD & 0.697(0.001) & 0.865(0.026) & 0.800(0.005) & 0.668(0.007) & 0.896(0.005) & \textbf{1.000(0.000)} & 0.564(0.026) \\
CHAMELEON & 0.706(0.004) & 0.727(0.019) & 0.669(0.008) & 0.651(0.010) & 0.758(0.016) & \textbf{1.000(0.000)} & 0.580(0.009) \\
\hline
\multicolumn{8}{|c|}{\textbf{\textit{Unit variance normalization}}} \\
\hline
\rowcolor{lgray}EOCC-1 & 0.693(0.005) & 0.867(0.005) & 0.853(0.020) & 0.670(0.024) & 0.928(0.011) & \textbf{1.000(0.000)} & 0.396(0.016) \\
\rowcolor{lgray}EOCC-2 & 0.831(0.001) & 0.878(0.006) & \textbf{0.995(0.001)} & \textbf{0.751(0.012)} & \textbf{0.957(0.004)} & \textbf{1.000(0.000)} & 0.460(0.026) \\
\rowcolor{lgray}EOCC-2\_10\% & 0.841(0.002) & 0.862(0.016) & \textbf{0.995(0.002)} & 0.709(0.023) & 0.954(0.007) & \textbf{1.000(0.000)} & 0.452(0.026) \\
\hline
Gauss & 0.862(0.000) & 0.899(0.005) & 0.985(0.001) & 0.721(0.003) & 0.929(0.003) & \textbf{1.000(0.000)} & 0.509(0.005) \\
MoG & 0.860(0.003) & 0.911(0.008) & 0.984(0.002) & 0.738(0.003) & 0.929(0.003) & \textbf{1.000(0.000)} & 0.494(0.006) \\
Na\"{\i}ve Parzen & 0.859(0.000) & \textbf{0.931(0.002)} & 0.987(0.001) & 0.678(0.003) & 0.930(0.008) & \textbf{1.000(0.000)} & 0.484(0.008) \\
Parzen & \textbf{0.877(0.001)} & 0.915(0.009) & 0.991(0.001) & 0.756(0.002) & 0.929(0.005) & \textbf{1.000(0.000)} & 0.469(0.008) \\
\textit{k}-Means & 0.801(0.003) & 0.902(0.009) & 0.984(0.001) & 0.712(0.010) & 0.878(0.015) & \textbf{1.000(0.000)} & 0.469(0.014) \\
1-NN & 0.862(0.001) & 0.914(0.012) & 0.991(0.001) & 0.721(0.002) & 0.906(0.008) & \textbf{1.000(0.000)} & 0.511(0.007) \\
\textit{k}-NN & 0.862(0.001) & 0.914(0.012) & 0.991(0.001) & 0.721(0.002) & 0.906(0.008) & \textbf{1.000(0.000)} & 0.511(0.007) \\
Auto-encoder & 0.836(0.000) & 0.890(0.013) & 0.960(0.002) & 0.658(0.005) &  0.888(0.023) & \textbf{1.000(0.000)} & \textbf{0.608(0.008)} \\
PCA & 0.826(0.001) & 0.776(0.031) & 0.920(0.004) & 0.640(0.006) & 0.655(0.013) & 0.920(0.008) & \textbf{0.608(0.008)} \\
SOM & 0.838(0.003) & 0.908(0.006) & 0.990(0.002) & 0.709(0.009) & 0.898(0.004) & \textbf{1.000(0.000)} & 0.487(0.017) \\
MST\_CD & - & 0.914(0.012) & 0.992(0.001) & 0.715(0.003) & 0.899(0.009) & \textbf{1.000(0.000)} & - \\
\textit{k}-Centres & 0.767(0.017) & 0.906(0.015) & 0.984(0.002) & 0.678(0.009) & 0.870(0.023) & \textbf{1.000(0.000)} & 0.483(0.006) \\
SVDD & 0.791(0.002) & 0.915(0.009) & 0.988(0.001) & 0.732(0.005) & 0.922(0.010) & \textbf{1.000(0.000)} & 0.490(0.010) \\
MPM & 0.735(0.002) & 0.909(0.010) & 0.991(0.001) & 0.729(0.003) & 0.922(0.007) & \textbf{1.000(0.000)} & 0.521(0.011) \\
LPDD & 0.751(0.002) & 0.889(0.008) & 0.989(0.001) & 0.634(0.005) & 0.947(0.004) & \textbf{1.000(0.000)} & 0.506(0.005) \\
CHAMELEON & - & - & - & - & - & - & - \\
\hline
\end{tabular}
\end{center}
\end{table*}
\begin{table*}[thp!]\scriptsize
\begin{center}
\caption{Same as Tab. \ref{tab:uci_ds_results}, but considering the high-dimensional UCI datasets.}
\label{tab:uci_ds_results2}
\begin{tabular}{|c|c|c|c|c|c|c|c|}
\hline
\textbf{System/Dataset} & \textbf{AR} & \textbf{BC-D} & \textbf{BC-P} & \textbf{C} & \textbf{CO} & \textbf{S} & \textbf{SP} \\
\hline
\multicolumn{8}{|c|}{\textbf{\textit{Non normalized data}}} \\
\hline
\rowcolor{lgray}EOCC-1 & 0.683(0.007) & 0.933(0.000) & 0.554(0.002) & 0.631(0.008) & 0.550(0.016) & 0.439(0.016) & 0.727(0.014) \\
\rowcolor{lgray}EOCC-2 & 0.775(0.016) & 0.938(0.001) & 0.585(0.021) & 0.654(0.014) & 0.783(0.032) & \textbf{0.998(0.000)} & 0.917(0.000) \\
\rowcolor{lgray}EOCC-2\_10\% & 0.707(0.019) & \textbf{0.946(0.005)} & 0.503(0.036) & 0.654(0.014) & 0.783(0.032) & 0.945(0.022) & 0.907(0.000) \\
\hline
Gauss & 0.606(0.006) & - & 0.591(0.009) & 0.704(0.011) & 0.803(0.017) & 0.680(0.031) & 0.833(0.033) \\
MoG & 0.577(0.166) & - & 0.511(0.017) & 0.500(0.000) & 0.500(0.011) & 0.704(0.035) & 0.776(0.031) \\
Na\"{\i}ve Parzen & 0.774(0.007) & - & 0.535(0.015) & 0.700(0.015) & 0.846(0.007) & 0.532(0.039) & 0.902(0.037) \\
Parzen & 0.577(0.166) & - & 0.586(0.029) & 0.364(0.224) & 0.502(0.022) & 0.805(0.031) & 0.879(0.027) \\
\textit{k}-Means & 0.766(0.006) & - & 0.536(0.021) & 0.668(0.031) & 0.862(0.025) & 0.698(0.037) & 0.923(0.017) \\
1-NN & 0.760(0.008) & - & 0.595(0.025) & 0.713(0.033) & 0.901(0.008) & 0.763(0.043) & 0.926(0.029) \\
\textit{k}-NN & 0.760(0.008) & - & 0.595(0.025) & 0.713(0.033) & 0.901(0.009) & 0.696(0.048) & 0.923(0.015) \\
Auto-encoder & 0.522(0.021) & - & 0.548(0.037) & 0.500(0.000) & 0.512(0.015) & 0.596(0.065) & 0.817(0.062) \\
PCA & \textbf{0.807(0.010)} & - & 0.574(0.018) & 0.707(0.016) & 0.824(0.004) & 0.696(0.033) & 0.901(0.030) \\
SOM & 0.772(0.007) & - & 0.523(0.030) & 0.682(0.026) & 0.887(0.020) & 0.801(0.034) & 0.975(0.021) \\
MST\_CD & 0.796(0.006) & - & \textbf{0.611(0.026)} & \textbf{0.733(0.030)} & \textbf{0.911(0.001)} & 0.811(0.031) & \textbf{0.981(0.026)} \\
\textit{k}-Centres & 0.767(0.016) & - & 0.584(0.055) & 0.684(0.029) & 0.815(0.036) & 0.668(0.041) & 0.909(0.016) \\
SVDD & 0.581(0.164) & - & 0.498(0.242) & 0.364(0.224) & 0.121(0.011) & 0.761(0.032) & \textbf{0.978(0.033)} \\
MPM & 0.771(0.005) & - & 0.053(0.001) & 0.500(0.000) & 0.901(0.006) & 0.785(0.030) & \textbf{0.980(0.074)} \\
LPDD & 0.577(0.166) & - & 0.539(0.183) & 0.418(0.200) & 0.864(0.004) & 0.636(0.027) & 0.934(0.033)  \\
CHAMELEON & 0.760(0.008) & - & - & 0.391(0.051) & 0.807(0.004) & 0.778(0.010) & 0.944(0.007) \\
\hline
\multicolumn{8}{|c|}{\textbf{\textit{Unit variance normalization}}} \\
\hline
\rowcolor{lgray}EOCC-1 & 0.657(0.035) & 0.843(0.026) & 0.492(0.021) & 0.452(0.010) & 0.550(0.016) & 0.291(0.026) & 0.664(0.014) \\
\rowcolor{lgray}EOCC-2 & 0.745(0.005) & \textbf{0.941(0.018)} & 0.490(0.040) & 0.506(0.009) & 0.783(0.032) & 0.349(0.019) & \textbf{0.959(0.002)} \\
\rowcolor{lgray}EOCC-2\_10\% & 0.707(0.019) & 0.922(0.010) & 0.503(0.016) & 0.506(0.009) & 0.783(0.032) & 0.357(0.027) & 0.927(0.017) \\
\hline
Gauss & 0.768(0.004) & - & 0.508(0.008) & 0.713(0.029) & 0.858(0.000) & 0.657(0.008) & 0.934(0.008) \\
MoG & 0.761(0.004) & - & 0.526(0.016) & - & - & 0.643(0.015) & 0.948(0.008) \\
Na\"{\i}ve Parzen & 0.774(0.007) & - & 0.538(0.022) & 0.700(0.015) & 0.846(0.000) & 0.569(0.012) & 0.892(0.009) \\
Parzen & 0.773(0.005) & - & 0.522(0.017) & 0.364(0.224) & 0.000(0.000) & 0.695(0.008) & 0.958(0.011) \\
\textit{k}-Means & \textbf{0.787(0.006)} & - & 0.520(0.020) & 0.716(0.040) & 0.872(0.005) & 0.625(0.016) & 0.867(0.010) \\
1-NN & 0.776(0.005) & - & 0.517(0.014) & \textbf{0.743(0.012)} & \textbf{0.888(0.000)} & 0.698(0.006) & \textbf{0.959(0.011)} \\
\textit{k}-NN & 0.776(0.005) & - & 0.517(0.014) & \textbf{0.743(0.012)} & \textbf{0.888(0.000)} & 0.698(0.006) & \textbf{0.959(0.011)} \\
Auto-encoder & - & - & 0.520(0.011) & - & - & 0.594(0.014) & 0.850(0.001) \\
PCA & 0.776(0.004) & - & \textbf{0.557(0.011)} & 0.707(0.019) & 0.853(0.000) & 0.608(0.009) & 0.807(0.020) \\
SOM & \textbf{0.787(0.008)} & - & 0.511(0.021) & 0.729(0.019) & 0.878(0.001) & 0.711(0.012) & 0.860(0.003) \\
MST\_CD & 0.778(0.005) & - & 0.527(0.018) & 0.735(0.022) & \textbf{0.888(0.000)} & \textbf{0.715(0.006)} & 0.957(0.011) \\
\textit{k}-Centres & 0.778(0.011) & - & 0.528(0.020) & 0.732(0.018) & 0.849(0.013) & 0.622(0.013) & 0.817(0.013) \\
SVDD & 0.527(0.094) & - & 0.517(0.017) & 0.364(0.224) & 0.000(0.000) & 0.705(0.054) & 0.897(0.032) \\
MPM & 0.771(0.005) & - & 0.518(0.018) & 0.000(0.000) & 0.324(0.000) & 0.696(0.008) & 0.901(0.028) \\
LPDD & 0.783(0.006) & - & 0.531(0.017) & 0.368(0.224) & 0.741(0.013) & 0.644(0.005) & 0.956(0.009) \\
CHAMELEON & - & - & - & - & - & - & - \\
\hline
\end{tabular}
\end{center}
\end{table*}

\subsection{Results on IAM Graph Datasets}
\label{sec:exp_iam}

The IAM repository is a variegated database containing many different datasets of labeled graphs \cite{riesen+bunke2008}.
We consider here six datasets, namely: AIDS, GREC, Letter-Low, Letter-High, Mutagenicity, and Protein. Tab. \ref{tab:iam_ds} shows the relevant information regarding the data. IAM datasets already contain a suitable validation set. Therefore, to adapt the considered dataset to our setting, we just move all non-target patterns of the training set into $\mathcal{S}_{ts}$.
We process the input labeled graphs by means of the graph edit distance algorithm known as TWEC \cite{odse,gm_survey}.
TWEC is a fast (quadratic) heuristic solution to the graph edit distance problem \cite{gm_survey}, which solves the assignment problem of the vertices by a greedy strategy. TWEC is characterized by three parameters ranging in $[0, 1]$, controlling the importance of insertion, deletion, and substitution edit operations.

In Tab. \ref{tab:details_results_iam} we show the obtained results. To our knowledge, there are no results available for comparison considering the one-class classification setting over the IAM datasets. However, we report those results to demonstrate the wide and straightforward applicability of EOCC and for future experimental comparisons.
The herein reported results confirm that EOCC-2 is more effective than EOCC-1, especially when considering harder datasets, such as M and P.
Notably, the P dataset is known to be very hard (see results in \cite{odse,odse2_ijcnn_2013} for the multi-class case); in Tab. \ref{tab:details_results_iam} the obtained AUC denotes nearly a randomized classifier.
However, the accuracy is fairly high ($\simeq 0.9$, with a precision of 1 and recall of $\simeq 0.21$). This proves that when considering the HDF, in this case the system correctly rejects all non-target patterns, while it rejects also some target instance.
In general, the gap between EOCC-1 and EOCC-2 is reduced for both AUC and accuracy.
The number of synthesized DRs is lower for EOCC-1. This aspect magnifies the potential of EOCC-1. In fact, in different experiments the results of EOCC-1 and EOCC-2 are comparable, although EOCC-1 solves the problem in a much lower computing time and with fewer DRs, i.e., with less resources.
This is an important aspect to be evaluated on the basis of the specific application at hand.
Standard deviations are acceptable, confirming the stability of the system.
\begin{table}[thp!]\scriptsize
\begin{center}
\caption{The considered IAM datasets. Full details in \cite{riesen+bunke2008}.}
\label{tab:iam_ds}
\begin{tabular}{|c|c|c|c|c|c|}
\hline
\textbf{IAM Dataset} & \textbf{Acronym} & \textbf{Target class} & \textbf{\# Target} & \textbf{\# Non-target} \\
\hline
AIDS & A & a & 200 & 1600 \\
\hline
GREC & G & 1 & 50 & 1033 \\
\hline
Letter-Low & L-L & A & 150 & 1400 \\
\hline
Letter-High & L-H & A & 150 & 1400 \\
\hline
Mutagenicity & M & mutagen & 2401 & 1713 \\
\hline
Protein & P & 1 & 99 & 332 \\
\hline
\end{tabular}
\end{center}
\end{table}
\begin{table*}[thp!]\scriptsize
\begin{center}
\caption{Test set results on IAM datasets. For each dataset, the first row shows results of EOCC-1, the second one those of EOCC-2.}
\label{tab:details_results_iam}
\begin{tabular}{|c|c|c|c|}
\hline
\textbf{Dataset} & \textbf{AUC} & \textbf{Accuracy} & \textbf{\# DRs} \\
\hline
\multirow{2}{*}{A} & 0.977(0.012) & 0.942(0.002) & 2.000(0.000) \\
%\cline{2-5}
& 0.974(0.014) & 0.945(0.012) & 3.400(2.607) \\
\hline
\multirow{2}{*}{G} & 0.993(0.006) & 0.969(0.008) & 4.000(0.707) \\
%\cline{2-5}
& 1.000(0.000) & 0.973(0.015) & 8.000(0.000) \\
\hline
\multirow{2}{*}{L-L} & 1.000(0.000) & 0.988(0.002) & 2.400(0.547) \\
%\cline{2-5}
& 1.000(0.000) & 0.990(0.001) & 3.800(0.447) \\
\hline
\multirow{2}{*}{L-H} & 0.905(0.116) & 0.840(0.144) & 2.000(0.000) \\
%\cline{2-5}
& 0.967(0.004) & 0.966(0.002) & 29.400(2.302) \\
\hline
\multirow{2}{*}{M} & 0.501(0.132) & 0.493(0.223) & 3.000(1.000) \\
%\cline{2-5}
 & 0.682(0.212) & 0.622(0.201) & 123.000(1.050) \\
\hline
\multirow{2}{*}{P} & 0.388(0.028) & 0.158(0.036) & 2.000(0.000) \\
%\cline{2-5}
& 0.554(0.025) & 0.921(0.012) & 26.000(1.732) \\
\hline
\end{tabular}
\end{center}
\end{table*}

%%%%%%%
\section{Conclusions and Future Directions}
\label{sec:conclusions}

In this paper, we have proposed and evaluated a novel one-class classification system, called EOCC.
The classifier has been designed by making use of an interplay of different techniques. The dissimilarity representation is exploited to make the system general-purpose. Graph-based techniques are employed for estimating information-theoretic quantities (i.e., the entropy) and for deriving the model of the classifier. The decision regions forming the model are obtained by exploiting the concept of modularity of a graph partition.
The decision regions are hence defined as clusters of vertices, which are further equipped with suitable membership functions.
This allows us to provide both hard (i.e., Boolean) and soft decisions about the recognition of test patterns.
We have validated the system over two types of benchmarks: (i) different feature-based UCI datasets and (ii) six IAM datasets of labeled graphs.
Overall, the comparisons made over the UCI datasets demonstrate the validity of the approach with respect to several state-of-the-art one-class classification systems taken from the literature.
Results on the IAM datasets prove the versatility and the effectiveness of the system in processing labeled graphs (a less conventional pattern type).

The one-class classification setting is very useful in all those situations where only patterns of interest are known. Such patterns are termed target instances.
Our solution is applicable to virtually any context, being based on the dissimilarity representation. This aspect is of particular interest, since it allows the user to model patterns according to their more suitable representation for the application at hand.

Future directions include usual improvements and variants of the herein discussed system (i.e., by changing suitable components but considering the same overall design). For instance, we used a complete Euclidean graph representation to estimate the entropy of the data mapped into the dissimilarity space. A more sparse representation could become handy when processing large volume of data.
Therefore, in the future we will evaluate entropic spanning graphs based on the so-called \textit{k}-NN graphs.
Other global optimization techniques are of course of interest, as well as other membership function models for the decision regions (e.g., by exploiting non-symmetric membership functions).
Particular attention will be devoted to the issue of \textit{interpretability} of the system.
Usually, researchers focus on improving performances of pattern recognition systems from the pure technical viewpoint, such as by paying attention on the generalization capability and the computing speed. However, when facing applications of true scientific interest, such systems should satisfy the requirement of producing results and using inference rules/functions that are easily understandable by humans.
This fact would allow field experts to easily gather useful insights about the underlying problem.
Our system is suitable for this mission, since it is conceived by exploiting fuzzy sets based techniques.
Therefore, an important future goal is the evaluation of the system from this specific viewpoint.

%\clearpage
\bibliographystyle{abbrvnat}
\bibliography{/home/lorenzo/University/Research/Publications/Bibliography.bib}
\end{document}